\documentclass[10pt,journal,compsoc]{IEEEtran}

\ifCLASSOPTIONcompsoc
\usepackage[nocompress]{cite}
\else
\usepackage{cite}
\fi
\usepackage{epsfig}
\usepackage{array}
\usepackage{graphicx}
\usepackage{amsmath,bm}
\usepackage{amssymb}
\usepackage{textcomp}
\usepackage{caption}
\usepackage{subcaption}
\usepackage[table,xcdraw]{xcolor}
\usepackage{multirow,booktabs,makecell}
\usepackage{setspace}
\usepackage{threeparttable}
\usepackage{float}
\usepackage{stfloats}
\usepackage{gensymb}

\usepackage{algorithm}
\usepackage{algorithmicx}
\usepackage{algpseudocode}
\usepackage{arydshln}
\usepackage{url}
\usepackage{soul}   
\soulregister\cite7
\soulregister\ref7
\usepackage{tabularx}
\usepackage{adjustbox}
\usepackage{enumitem}

\usepackage{calc}
\newlength\myheight
\newlength\mydepth
\settototalheight\myheight{Xygp}
\newcommand*\inlinegraphics[1]{%
  \settototalheight\myheight{Xygp}%
  \settodepth\mydepth{Xygp}%
  \raisebox{-\mydepth}{\includegraphics[height=\myheight]{#1}}%
}

\newcommand{\etal}{\textit{et al.}}
\newcommand{\ie}{\textit{i.e.}}
\newcommand{\eg}{\textit{e.g.}}
\newcommand{\wrt}{\textit{w.r.t. }}
\newcommand{\nbf}[1]{{\noindent \textbf{#1}}}

\begin{document}
%
\title{ETPNav: Evolving Topological Planning for Vision-Language Navigation in\\ Continuous Environments}
\author{Dong An, 
        Hanqing Wang,
        Wenguan Wang,
        Zun Wang,
        Yan Huang,
        Keji He,
        Liang Wang
        
    \IEEEcompsocitemizethanks{
        \IEEEcompsocthanksitem 
        D. An, Y. Huang, K. He and L. Wang are with the Center for Research on Intelligent Perception and Computing (CRIPAC), National Laboratory of Pattern Recognition (NLPR); School of Future Technology and School of Artificial Intelligence, University of Chinese Academy of Sciences, China. Email: 
        dong.an@cripac.ia.ac.cn, yhuang@nlpr.ia.ac.cn, keji.he@cripac.ia.ac.cn, wangliang@nlpr.ia.ac.cn.
        \IEEEcompsocthanksitem 
        H. Wang is with the Beijing Institute of Technology, China. Email: hanqingwang@bit.edu.cn.
        \IEEEcompsocthanksitem 
        W. Wang is with the Zhejiang University, China. Email: wenguanwang.ai@gmail.com. 
        \IEEEcompsocthanksitem 
        Z. Wang is with the Australian National University. Email: zun.wang@anu.edu.au.




        
        \IEEEcompsocthanksitem
        Wenguan Wang and Yan Huang are the corresponding authors.
}}
\markboth{Journal of \LaTeX\ Class Files,~Vol.~*, No.~*, *~*}%
{Shell \MakeLowercase{\textit{et al.}}: ETPNav}

\IEEEtitleabstractindextext{%
\begin{abstract}
Vision-language navigation is a task that requires an agent to follow instructions to navigate in environments. It becomes increasingly crucial in the field of embodied AI, with potential applications in autonomous navigation, search and rescue, and human-robot interaction.
In this paper, we propose to address a more practical yet challenging counterpart setting - vision-language navigation in continuous environments (VLN-CE). 
To develop a robust VLN-CE agent, we propose a new navigation framework, ETPNav, which focuses on two critical skills: 1) the capability to abstract environments and generate long-range navigation plans, and 2) the ability of obstacle-avoiding control in continuous environments.
ETPNav performs online topological mapping of environments by self-organizing predicted waypoints along a traversed path, without prior environmental experience. It privileges the agent to break down the navigation procedure into high-level planning and low-level control.
Concurrently, ETPNav utilizes a transformer-based cross-modal planner to generate navigation plans based on topological maps and instructions.
The plan is then performed through an obstacle-avoiding controller that leverages a trial-and-error heuristic to prevent navigation from getting stuck in obstacles. 
Experimental results demonstrate the effectiveness of the proposed method. ETPNav yields more than 10\% and 20\% improvements over prior state-of-the-art on R2R-CE and RxR-CE datasets, respectively.
 Our code is available at \url{https://github.com/MarSaKi/ETPNav}.
\end{abstract}

}
\maketitle

\IEEEdisplaynontitleabstractindextext
\IEEEpeerreviewmaketitle

\IEEEraisesectionheading{\section{Introduction}\label{sec:introduction}}

\IEEEPARstart{G}iven a natural language instruction, the task of vision-language navigation (VLN)~\cite{anderson2018vision} requires an agent to interpret and follow the instruction to reach the target location. 
This task has been well-studied over the past few years~\cite{wang2019reinforced,wang2021structured,hong2021vln,chen2021history}, however, the majority of works focus on the discrete VLN setting. 
This setting simplifies navigation as traversing on a predefined graph of an environment, which significantly narrows down the possible locations of the agent and target.
Recognizing this cannot reflect the challenges of a deployed system encountered in a real environment, Krantz \etal~\cite{krantz2020beyond} introduce VLN in continuous environments (VLN-CE), which discards the strong graph assumption, instead, requires the agent to navigate on a 3D mesh freely with low-level actions.

So far, VLN-CE has been shown far more difficult than VLN, with a few published works revealing episode success rates less than half of those reported in VLN. 
Early efforts for VLN-CE are end-to-end trained systems to directly predict low-level actions (or waypoints) from language and observations~\cite{krantz2020beyond,raychaudhuri2021language,krantz2021waypoint}. This scheme can be challenged by joint learning of navigation and language grounding in a long-horizon task, thus leading to a lower performance compared to VLN. 
Recently, there has been an emerging trend towards modular waypoint-based approaches~\cite{hong2022bridging,krantz2022sim,an20221st} that divide the complex task into waypoint generation, subgoal planning, and navigation control. 
Concretely, in each decision loop, the agent uses a pre-trained network to predict several nearby candidate waypoints, and then performs cross-modal grounding to select a subgoal from the waypoints. After that, a controller drives the agent to reach the selected subgoal with low-level actions. 
Overall, this modular pipeline simplifies policy learning and closes the performance gap between VLN-CE and VLN.

Despite the progress, we find these waypoint-based methods still have drawbacks in three aspects. 
\textbf{First}, the predicted waypoints are still local and constrained to a nearby area of the agent, which are insufficient to capture the global environment layouts and may hinder the agent's long-range planning capacity. For example, to backtrack to a previous remote location for past decision correction, the agent has to run multiple plan-control flows which can introduce unstable accumulation bias. 
\textbf{Second}, the key design choices for waypoint prediction have not been well-studied. One representative predictor~\cite{hong2022bridging} takes RGBD images as inputs, but whether or not the semantic-level RGB inputs are valid remains unknown, since it is only tasked with inferring spatial accessibility.
\textbf{Third}, different types of controllers (heuristic~\cite{krantz2021waypoint,hong2022bridging}, map-based~\cite{sethian1996fast,krantz2022sim}, learning-based~\cite{wijmans2019dd,georgakis2022cross}) are proposed for waypoint-reaching, but their robustness to obstacles remains unknown. 
As a result, an obstacle-intolerant controller can cause the agent to get stuck in obstacles, leading to navigation failure.

To address the above problems, we propose a hierarchical navigation framework powered by topological (topo) maps, and a low-level controller. 
Topo maps, partially inspired by cognitive science~\cite{udin1988formation}, typically depict environments as low-dimensional graph representations with nodes for places and edges for reachability. 
They can efficiently capture environment layouts and long-range navigation dependency, thereby easing the agent to make long-range goal plans, such as planning a shortest path within the map to reach a remote location. 
But what makes our topo maps novel is that they are constructed via online self-organization of predicted waypoints, which are concise and meet the assumption of partial observability in a real environment. 
Notably, this scheme is greatly distinct from previous VLN literature regarding topo mapping, which requires either predefined graphs~\cite{deng2020evolving,wang2021structured,chen2022think} or environment pre-exploration~\cite{chen2021topological}.

To better capture the environment layouts, we systematically examine our topo map's key design choices, such as waypoint prediction, node density, and node representation. In particular, we find that a depth-only waypoint predictor aids generalization in novel environments, while RGB information may undermine spatial accessibility inferring.
Moreover, we explicitly consider the obstacle-avoidance problem in VLN-CE. We find this problem is especially crucial in a more challenging and practical scenario - sliding along obstacles is forbidden, where commonly used controllers~\cite{krantz2021waypoint,wijmans2019dd,sethian1996fast} can cause navigation to get stuck in obstacles frequently, resulting in a severe performance drop. Accordingly, we propose a new controller via a trial-and-error heuristic to explicitly help the agent escape from deadlocks, nearly eliminating the performance loss caused by sliding-forbidden.

Altogether, we propose a full navigation system for VLN-CE. 
For each episode, our agent updates a topo map through online self-organization of waypoints predicted so far. 
The map decomposes the navigation problem into planning and control. Within each decision loop, the agent uses a cross-modal transformer~\cite{vaswani2017attention,xu2023multimodal} to compute a global navigation plan from the instruction and the topo map. Then, this plan is executed by a robust obstacle-avoiding controller with low-level actions.

Extensive experiments demonstrate the effectiveness of the proposed method, and our system achieves state-of-the-art on two VLN-CE benchmarks (\eg, in test unseen splits, 55 SR and 48 SPL on R2R-CE dataset, 51.21 SR and 41.30 SDTW on RxR-CE dataset). 
Based on the algorithm described in this paper, we won the CVPR 2022 RxR-Habitat Challenge~\cite{deitke2022retrospectives,an20221st}.
In summary, the contributions of this work are four-fold:

\begin{itemize}[leftmargin=*]
\setlength{\itemsep}{0pt}
\setlength{\parsep}{-0pt}
\setlength{\parskip}{-0pt}
\setlength{\leftmargin}{-8pt}
\item We propose a new topological map-based method for robust navigation planning in VLN-CE. It can efficiently abstract the continuous environments and facilitates the agent's long-range goal planning.
\item We investigate the essential design choices for building topological maps through comprehensive experiments, demonstrating that a concise depth-only design is optimal for waypoint prediction.
\item We study a practically important but rarely investigated problem in VLN-CE - obstacle avoidance, and propose an effective trial-and-error controller to address the problem. 
\item The proposed system won the CVPR 2022 RxR-Habitat Challenge and doubled the SDTW of the second-best model. It can serve as a strong baseline for further research on this challenging task.\footnote{Our code is available at \url{https://github.com/MarSaKi/ETPNav}.} 
\end{itemize}

The rest of this paper is organized as follows. In \S~\ref{sec:related}, we give a brief review of the related work. \S~\ref{sec:method} describes the task setup of vision-language navigation in continuous environments and then introduces our proposed method. Experimental results are provided in \S~\ref{sec:exprs}. Lastly, we conclude this work in \S~\ref{sec:conclusion}.

\section{Related Work}\label{sec:related}
\subsection{Vision-Language Navigation}
Learning navigation with language guidance has drawn significant research interest in recent years. 
R2R~\cite{anderson2018vision} and RxR~\cite{ku2020room} datasets introduce low-level human language instructions and photo-realistic environments for indoor navigation, while Touchdown~\cite{chen2019touchdown} further extends this task in an outdoor navigation context. 
Following these works, dialogue-based navigation such as CVDN~\cite{thomason2020vision} and HANNA~\cite{nguyen2019help}, and navigation for remote object-finding such as REVERIE~\cite{qi2020reverie} and SOON~\cite{zhu2021soon} have been proposed for further research. 

Early VLN methods use sequence-to-sequence LSTMs to predict low-level actions~\cite{anderson2018vision} or high-level actions from panoramas~\cite{fried2018speaker}. 
Various attention mechanisms~\cite{ma2019self,qi2020object,hong2020language,an2021neighbor} are proposed to improve the learning of visual-textual correspondence. Reinforcement learning is also explored to improve policy learning~\cite{wang2018look,wang2019reinforced,tan2019learning,wang2020soft}.
Different strategies are also investigated to form a more robust navigation policy, such as environment pre-exploration~\cite{wang2019reinforced}, active perception~\cite{wang2020active,wang2023active}, and planning with graph memory~\cite{deng2020evolving,wang2021structured}. 
To enhance an agent's generalization ability to novel environments, various data augmentation strategies are studied to mimic new environments~\cite{tan2019learning,liu2021vision,li2022envedit,parvaneh2020counterfactual,kamath2022new} or synthesis new instructions~\cite{fried2018speaker,wang2022counterfactual,wang2022less,zhao2021evaluation,wang2023lana,wang2023learning}. 
Recently, transformer-based models have shown superior performance thanks to their powerful ability to learn generic multi-modal representations~\cite{hao2020towards,guhur2021airbert,majumdar2020improving}. This scheme is further extended by recurrent agent state~\cite{hong2021vln,qi2021road,moudgil2021soat}, episodic memory~\cite{chen2021history,qiao2023hop+,lin2022multimodal}, graph memory~\cite{zhao2022target,chen2022think,hwang2023meta,wang2023BEV,an2022bevbert} and prompt learning~\cite{lin2022adapt,liang2022visual} that significantly improves sequential action prediction. 

Despite the progress, these agents are developed under the discrete VLN setting, which simplifies navigation as traversing on a predefined graph of an environment. 
In effect, this setup greatly narrows down the possible locations of the agent and target, while ignoring the low-level control problem that arises in a real-world navigation system. As a result, directly transferring these agents into the real world~\cite{anderson2021sim} or continuous environments~\cite{krantz2020beyond} can cause a severe performance drop.

\vspace{-2mm}
\subsection{VLN in Continuous Environments}
Recognizing the navigation graph assumption cannot reflect the challenges a deployed system would experience in a real environment, Krantz \etal~\cite{krantz2020beyond} introduce VLN in continuous environments (VLN-CE) - requiring the agent to navigate on a 3D mesh freely with low-level actions (\eg, \textrm{FORWARD} 0.25m, \textrm{ROTATE} 15\degree). 
To benchmark VLN-CE agents, discrete paths in R2R~\cite{anderson2018vision} and RxR~\cite{ku2020room} are transferred to continuous environments through the Habitat Simulator~\cite{savva2019habitat}.

Initial methods for VLN-CE are end-to-end trained systems to directly predict low-level actions from language and observations~\cite{krantz2020beyond,raychaudhuri2021language,irshad2021sasra}, but demonstrate a huge performance gap to VLN.
Because jointly learning language-grounding and low-level control in a fully end-to-end manner can be difficult and expensive, typically requiring millions of frames of experience~\cite{wijmans2019dd}. 
Thus, Krantz \etal~\cite{krantz2021waypoint} propose to decouple the navigation process as subgoal planning and low-level control, where the model predicts a language-conditioned waypoint as subgoal at each step, and then reaches the subgoal with 
a rotate-then-forward control flow.
This idea is further extended with semantic maps for better perception of environments~\cite{georgakis2022cross,chen2022weakly}, finding increased performance but still far below VLN. 
One potential reason is that the prediction of language-conditioned waypoints requires the model to learn spatial accessibility inferring and cross-modal reasoning simultaneously, which is difficult and may need massive training~\cite{krantz2021waypoint}.

Recently, there has been an emerging trend towards modular waypoint-based approaches~\cite{krantz2022sim,hong2022bridging,an20221st,wang2023dreamwalker}. Instead of directly predicting a language-conditioned waypoint, these methods further decouple the subgoal prediction as candidate waypoint generation and subgoal selection. Concretely, within each decision loop, the agent first uses a pre-trained network to predict several nearby candidate waypoints, and then performs cross-modal grounding over the waypoints to select a subgoal. Similarly, the subgoal-reaching is conducted by a follow-up controller. 
Overall, this modular scheme simplifies policy learning and further closes the gap between VLN-CE and VLN. 
But the drawback is the local waypoint representations which are insufficient to capture global environment layouts and navigation dependency, leading to the agent's non-ideal long-range planning capability. Meanwhile, the widely used controllers are unaware of obstacles, and we find they can cause navigation to get stuck in obstacles frequently in a practical sliding-forbidden scenario. 
To address these limitations, we not only propose an online constructed topo map for long-range planning, but also devise an obstacle-avoiding controller.

\begin{figure*}[!htbp]
\centering
\includegraphics[width=0.96\textwidth]{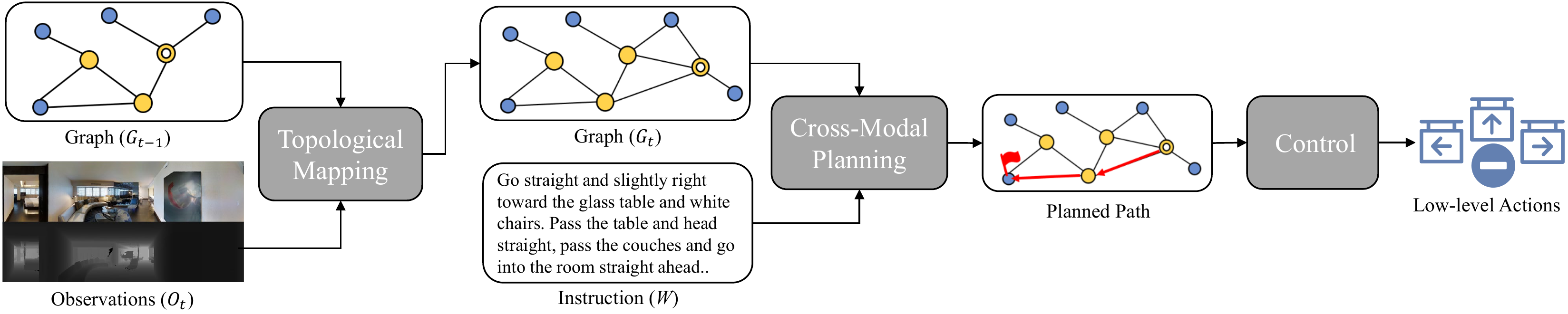}
\caption{
Overview of the proposed model, ETPNav. 
It consists of three modules, a topological mapping module that gradually updates the topological map as it receives new observations, a cross-modal planning module that computes a navigational plan based on the instruction and map, and a control module that executes the plan with low-level actions.
}\label{fig:overview}
\end{figure*}
\vspace{-2mm}

\subsection{Maps for Navigation}
Works on robot navigation have a long tradition of using spatial or topological space representations to aid environmental perception. 
Researchers have investigated explicit metric spatial representations~\cite{elfes1989using} and their construction using various sensors~\cite{mur2015orb,newcombe2011dtam}, as well as agent localization with such representations~\cite{dellaert1999monte,henriques2018mapnet}.
Modern literature has begun to integrate spatial representations with semantics, yielding promising results in tasks such as active perception~\cite{chaplot2021seal,chaplot2019learning}, object-goal navigation~\cite{chaplot2020object,gervet2022navigating}, vision-language navigation~\cite{georgakis2022cross,chen2022think,min2021film} and audio-visual navigation~\cite{chen2023omnidirectional}. 
However, these metric maps typically suffer from scalability issues, rendering them unsuitable for long-range navigation tasks such as VLN-CE. 
For instance, though a large metric map has been shown effective in VLN-CE, it may result in excessive computation costs~\cite{georgakis2022cross}. 
To that end, non-metric topological representations have also been considered in classical literature~\cite{choset2001topological,kuipers1991robot}, and researchers have investigated the use of semantic topo maps for high-level navigation tasks~\cite{chaplot2020neural,kwon2021visual,hahn2021no}. 
Topo maps, based on low-dimensional graph representations, efficiently capture environment layouts and are beneficial for exploration or long-range planning. 

In VLN, several works have employed topo maps and demonstrated superior performance~\cite{deng2020evolving,wang2021structured,chen2022think,zhao2022target}. Because the long-range map can facilitate the agent to learn self-correction policy, which is crucial when the agent loses track of an instruction. 
However, these maps are derived from predefined graphs by marking observed nodes, which are unavailable in continuous or real-world environments.  
Chen \etal~\cite{chen2021topological} explored topo maps in VLN-CE, but their proposed map is built offline through environment pre-exploration and assumes the agent has access to global topology priors, which limits its use in more realistic scenarios. 
Inspired by their novel ideas, we propose a more practical solution for topo mapping in VLN-CE.
Without the need for predefined graphs or environment pre-exploration, our map is built online through the self-organization of predefined waypoints at each step. It is scalable as navigation progresses and meets the assumption of partial observability in a real environment.

\section{Method}\label{sec:method}
\nbf{Task Setup.} 
We address instruction-following navigation in indoor environments, where an agent is required to follow a specific path described by a natural language instruction to reach the target location. 
In particular, we focus on a practical setup - vision-language navigation in continuous environments (VLN-CE)~\cite{krantz2020beyond}, where the agent navigates on a 3D mesh of an environment with low-level actions. The action space consists of a set of parameterized discrete actions (\eg, \textrm{FORWARD} (0.25m), \textrm{ROTATE LEFT/RIGHT} (15\degree), and \textrm{STOP}). 
VLN-CE uses the Habitat Simulator~\cite{savva2019habitat} to render environmental observations based on the Matterport3D scene dataset~\cite{chang2017matterport3d}. 
Following the panoramic VLN-CE setting~\cite{krantz2021waypoint,hong2022bridging,krantz2022sim}, at each location, the agent receives panoramic RGB observations $O=\{I^{rgb}, I^{d}\}$ consisting of 12 RGB images and 12 depth images, which are captured from different views at 12 equally-spaced horizontal heading angles, \ie, $( 0\degree, 30\degree, ..., 330\degree )$. 
The agent also receives an instruction for each episode. We denote the embeddings of the instruction with $L$ words by $W=\{\bm{w}_i\}_{i=1}^L$.

\vspace{2mm}
\nbf{Overview of Our Approach.}
We propose a hierarchical navigation model, named `ETPNav', which leverages high-level topological map-based planning and low-level controller for the VLN-CE task.
As illustrated in Fig.~\ref{fig:overview}, ETPNav comprises three modules: topological mapping, cross-modal planning and control. 
%
%
The mapping module maintains a topo map for each episode. 
Within each decision loop, the mapping module first updates the topo map by incorporate current observations. 
Subsequently, the planning module conducts cross-modal reasoning over the map and instruction to craft a high-level topological path plan. 
The plan is then executed by the control module, which drives the agent through a sequence of low-level actions. 
To be noted, in the following technique descriptions, we use `step $t$' to denote the decision loop step, rather than the low-level action step. 

Similar to recent work~\cite{georgakis2022cross,chen2022weakly,krantz2022sim}, we presume that the agent can access the ground-truth pose provided by the simulator to facilitate mapping and control. 
Note that this work does not address the challenge of estimating pose based on noisy sensor readings. However, we suggest that visual odometry techniques~\cite{zhao2021surprising} may be adaptable to our model in this context. 
This paper proceeds by introducing topological mapping in \S~\ref{sec:mapping}, followed by cross-modal planning in \S~\ref{sec:planning}, and the presentation of our control policy in \S~\ref{sec:control}. Finally, we provide detailed expositions of training and inference of our model in \S~\ref{sec:training}.


\subsection{Topological Mapping}\label{sec:mapping}
\begin{figure*}[!htbp]
\centering
\includegraphics[width=0.96\textwidth]{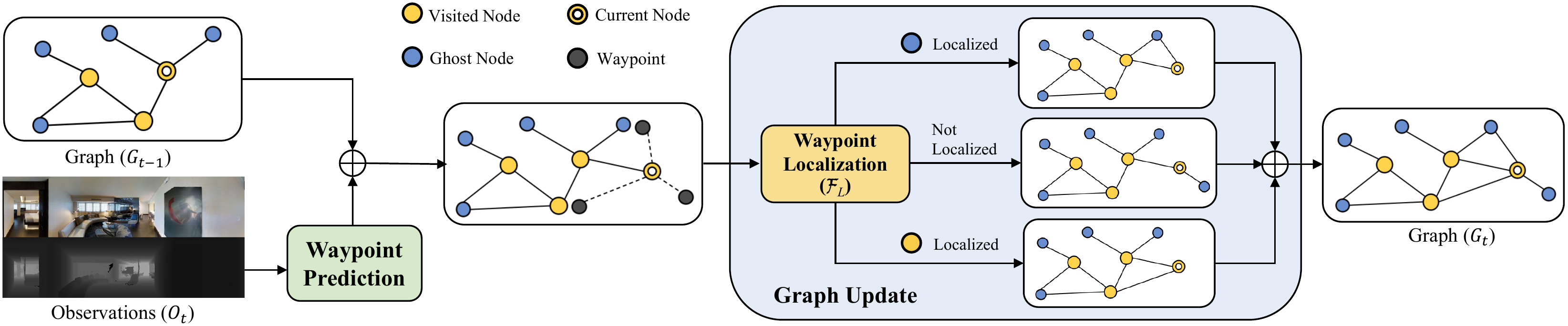}
\vspace{-1mm}
\caption{
Illustration of the topological mapping module. It takes the previous graph ($G_{t-1}$) and the agent observation ($O_t$) as input. The waypoint prediction submodule first predicts several nearby waypoints. The graph update submodule organizes these waypoints and incorporates them to update the graph using a waypoint localization function ($\mathcal{F}_L$).
}\label{fig:mapping}
\vspace{-2mm}
\end{figure*}
To facilitate long-range planning, our agent constructs a topo map on the fly. 
This map shares a similar structure with~\cite{chaplot2020neural}, which abstracts the visited or observed locations along the traversed path as a graph representation, denoted as $G_t=\langle N_t, E_t\rangle$ at step $t$.
Each node ($n_i \in N_t$) contains visual information observed at its location as well as position information.  
Two nodes are connected by an edge ($e_{i,j} \in E_t$) if their represented locations are directly reachable from each other. Each edge also stores the relative Euclidean distance between two nodes. 
We divide these nodes into visited nodes~\inlinegraphics{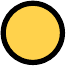}, the current node~\inlinegraphics{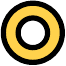}, and ghost nodes~\inlinegraphics{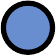}, where `ghost' denotes that nodes have been observed but left unexplored. 

Different from prior work~\cite{deng2020evolving,wang2021structured,chen2022think,chen2021topological}, our method assumes no prior knowledge of the environmental structure and we propose to construct the topo map via online self-organization of predicted waypoints. 
As depicted in Fig.~\ref{fig:mapping}, at each step $t$, the agent first predicts several nearby waypoints~\inlinegraphics{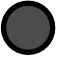}, representing possibly accessible locations near the agent. 
A current node~\inlinegraphics{fig/current_node.pdf} is also initialized at the agent's current location and connects to the last visited node (if it exists). 
The predicted waypoints and current node are represented by feature embeddings of the current observations $O_t$. 
These waypoints will be organized to update the previous topo map $G_{t-1}$ and obtain the current map $G_{t}$.

\vspace{1mm}
\nbf{Image Processing.}
Given the current step's RGBD observations $O_t=\{I_{t}^\text{rgb}, I_{t}^\text{d}\}$, two different pre-trained visual encoders are used to extract RGB feature vectors $\bm{V}_{t}^\text{rgb}\!=\!\{ \bm{v}_{i}^\text{rgb} \}_{i=1}^{12}$ and depth feature vectors $\bm{V}_{t}^\text{d}=\{ \bm{v}_{i}^\text{d} \}_{i=1}^{12}$, respectively. 
To distinguish the features captured from different views of the panorama, we also apply orientation features $\bm{V}_{t}^\text{ori}=\{(\cos \theta_i, \sin \theta_i)\}_{i=1}^{12}$, where $\theta_i$ represents heading angle. The parameters of the two visual encoders are fixed. More details of pre-processing are introduced in \S~\ref{sec:imp}.

\vspace{1mm}
\nbf{Waypoint Prediction.}
We employ a transformer-based waypoint predictor~\cite{hong2022bridging} to generate the nearby waypoints. 
The predictor takes the depth feature vectors $V_{t}^\text{d}$ and orientation feature vectors $V_{t}^\text{ori}$ to predict the relative poses of these waypoints. 
Concretely, feature vectors in $V_{t}^\text{d}$ and $V_{t}^\text{ori}$ are first fused using a linear layer. 
The resulting vectors are fed into a two-layer transformer to conduct inter-view interaction and obtain contextual depth embeddings. 
These embeddings are then fed into a multi-layer perceptron to obtain a heatmap representing probabilities of nearby waypoints in space. 
$K$ waypoints $\triangle P^\text{w} = \{\triangle p_{i}^\text{w}\}_{i=1}^K$ are sampled from the heatmap using a non-maximum-suppression (NMS), where $\triangle p_{i}^\text{w}$ denotes the relative pose to the agent. 
The predictor is pre-trained on the MP3D graph dataset~\cite{hong2022bridging}, and its parameters are fixed. 

To be noted, our predictor only takes depth images as inputs, instead of RGBD images used in~\cite{hong2022bridging}. Such depth-only design is motivated by the fact that waypoints only represent spatial accessibility, while semantic-level RGB information may be not helpful or even detrimental. We provide an ablation analysis of this design in \S~\ref{sec:ab_map}.

\vspace{1mm}
\nbf{Visual Representations for Waypoints and the Current Node.}
We conduct feature mapping of the current observations $O_t$ to represent the predicted waypoints and the current node. 
Specifically, RGB features $\bm{V}_{t}^\text{rgb}$, depth features $\bm{V}_{t}^\text{d}$ and orientation features $\bm{V}_{t}^\text{ori}$ are fused using a linear layer, and then fed into a panorama encoder.
The panorama encoder uses a multi-layer transformer to perform inter-view interaction and outputs contextual visual embeddings $\widehat{\bm{V}}_{t}=\{\hat{v}_i\}_{i=1}^{12}$.
The current node~\inlinegraphics{fig/current_node.pdf} has access to the panoramic observations and thus is represented as an average of $\widehat{\bm{V}}_{t}$.
The waypoints~\inlinegraphics{fig/waypoint.pdf} are partially observed and are represented by embeddings of views from which they can be observed. For example, if the relative heading angle of a waypoint to the agent is within $0\degree \sim 30\degree$, the waypoint is represented by the first view embedding $\hat{\bm{v}}_{1}$. The waypoint representations will be incorporated to update the representations of ghost nodes~\inlinegraphics{fig/ghost_node.pdf}.

\vspace{1mm}
\nbf{Graph Update.}
We update the topo map with the predicted waypoints based on their spatial relations with existing nodes in the graph. 
This process utilizes a Waypoint Localization ($\mathcal{F}_L$) function to localize waypoints in the graph.
$\mathcal{F}_L$ takes the position of a waypoint as input and computes its Euclidean distances with all nodes in the graph. 
If the minimum distance is less than a threshold $\gamma$, $\mathcal{F}_L$ returns the corresponding node as the localized node.  
For each waypoint, we try to localize it in the graph using the Waypoint Localization function ($\mathcal{F}_L$). 
To update the graph, we divide the localization results into three cases:
\begin{enumerate}[leftmargin=*]
\item If a visited node~\inlinegraphics{fig/visited_node.pdf} is localized, delete the input waypoint and add an edge between the current node and the localized visited node. 
\item If a ghost node~\inlinegraphics{fig/ghost_node.pdf} is localized, accumulate the position and visual representation of the input waypoint to the localized ghost node. The new position and representation of the localized ghost node are updated as the average of its accumulated waypoint positions and representations.
\item If no node is localized, we take the input waypoint as a new ghost node.
\end{enumerate}

\subsection{Cross-Modal Planning}\label{sec:planning}
\begin{figure}[t]
\centering
\includegraphics[width=0.49\textwidth]{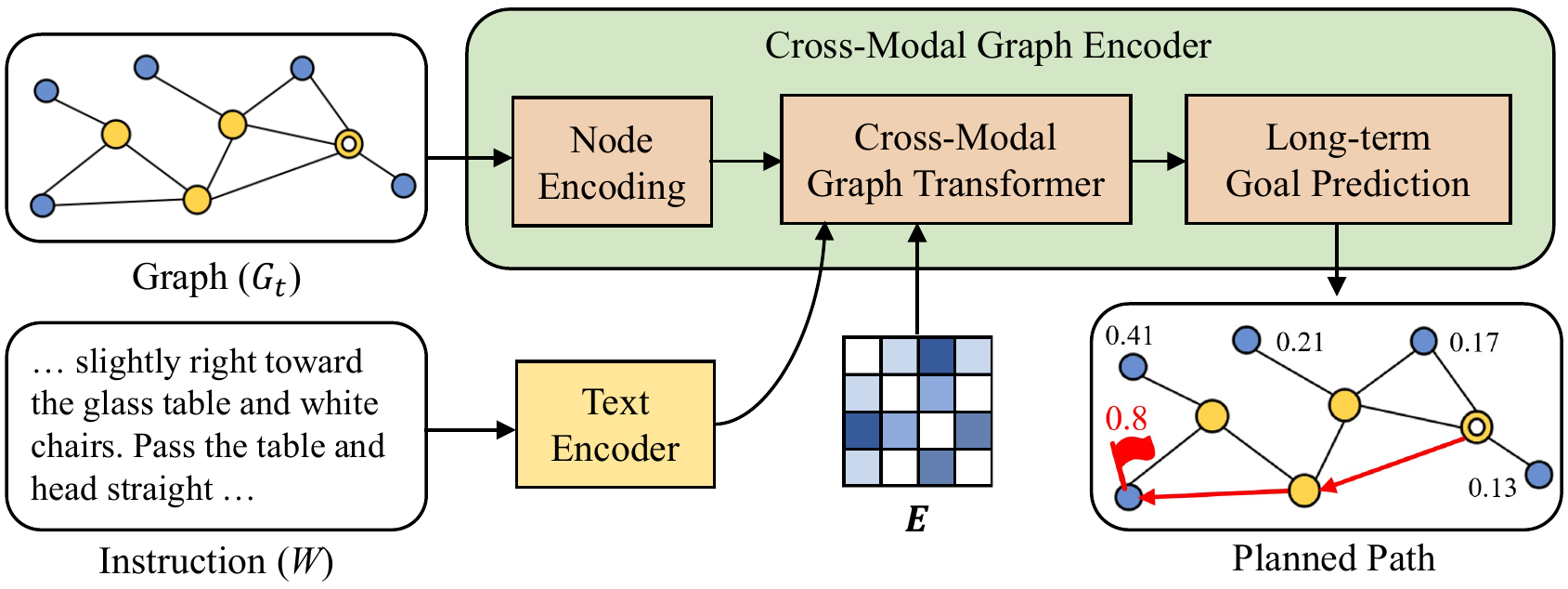}
\vspace{-1mm}
\caption{
The planning module consists of a text encoder for instruction encoding, 
and a graph encoder to conduct cross-modal reasoning over the map to generate a path plan.
}\label{fig:planning}
\vspace{-2mm}
\end{figure}

Fig.~\ref{fig:planning} illustrates the cross-modal planning module. 
It consists of a text encoder and a cross-modal graph encoder. 
The instruction of the current episode is encoded by the text encoder.
Then, the cross-modal graph encoder conducts reasoning over the topo map and encoded instruction to predict a long-term goal node. 
The output is a planned topological path to the goal. 

\subsubsection{Text Encoder}
Each word embedding $\bm{w}_i$ is added a positional embedding~\cite{kenton2019bert} corresponding to the position of the word in the sentence and a type embedding for text~\cite{tan2019lxmert}. We denote the word embeddings with positional information as $\widehat{W}=\{\hat{\bm{w}}_i\}_{i=1}^L$. 
Those embeddings are then fed into a multi-layer transformer to obtain contextual word representations.

\vspace{-3mm}
\subsubsection{Cross-Modal Graph Encoder}
The module takes the topo map $G_t$ and encoded instruction $\widehat{W}$ to predict a long-term goal node in the topo map. 

\vspace{1mm}
\nbf{Node Encoding.}
The visual feature in node $n_i$ is added with a pose encoding and a navigation step encoding. 
The pose encoding embeds the global relative pose information of a node \wrt the agent’s current location, including its orientation and Euclidean distance relative to the current node. 
The navigation step encoding embeds the latest visited time step for visited nodes and 0 for ghost nodes. 
This allows visited nodes to be encoded with different histories to capture navigation dependencies and facilitate alignment with the instruction. 
The encoding of $n_i$ is denoted as $\bm{n}_i$. 
To represent a \textrm{STOP} action, we add a `stop' node in the graph and connect it with all other nodes. 

\vspace{1mm}
\nbf{Cross-Modal Graph Transformer.}
The encoded node and word embeddings are fed into a multi-layer transformer to conduct cross-modal interaction.
The transformer architecture is similar to LXMERT~\cite{tan2019lxmert}, with each layer comprising one bi-directional cross-attention sub-layer, two self-attention sub-layers, and two feed-forward sub-layers.
For node encoding, the standard self-attention layer~\cite{vaswani2017attention} only considers visual similarity among nodes, which may overlook nearby nodes which are more relevant than distant nodes.
To this end, we devise a graph-aware self-attention (GASA) that further takes into account the graph topology when computing inter-node attention for node encoding:
\begin{equation}
\textrm{GASA}(\bm{X})=\textrm{softmax}(\frac{\bm{X} \bm{W}_\text{q} (\bm{X} \bm{W}_\text{k})^\top}{\sqrt{d}} 
+ \bm{E}\bm{W}_\text{e})) \bm{X} \bm{W}_\text{v}
\label{eq:gasa}
\end{equation}
where $\bm{X}$ represents the stack of all node encodings, $\bm{E}$ is the spatial matrix constructed by all-pair shortest distances obtained from the graph edges $E_t$, $\bm{W}_\text{q},\bm{W}_\text{k},\bm{W}_\text{e},\bm{W}_\text{v}$ are learnable matrices. The produced visual-textual associated representation of nodes is formulated as 
$[ \tilde{\bm{n}}_1, \hdots, \tilde{\bm{n}}_{|N_t|} ]=\textrm{GASA}([ \bm{n}_1, \hdots, \bm{n}_{|N_t|} ])$.

\vspace{1mm}
\nbf{Long-term Goal Prediction.}
We predict a navigation goal score for each node in the topo map $G_t$ as follows:
\begin{equation}
\bm{s}_i = \textrm{FFN} (\tilde{\bm{n}}_i)
\label{eq:predict}
\end{equation}
where \textrm{FFN} denotes a feed-forward network and $\tilde{\bm{n}}_i$ is the multimodal representation of node $n_i$. 
Note that $s_0$ corresponds to the `stop' node and it represents the score of the \textrm{STOP} action. 
To avoid unnecessary repeated visits to visited nodes, we mask the score for visited nodes and the current node. 
As such, a long-term goal is picked from ghost nodes or the `stop' node.

Finally, the agent selects a long-term goal according to the predicted goal scores (\eg, pick the node with the maximum score). 
If the selected goal is the `stop' node, navigation of the current episode terminates. 
If the selected goal is a ghost node, the agent computes a shortest path to the goal by performing Dikjstra's algorithm on the graph. The resulting path plan consists of a sequence of subgoal nodes, denoted as $\mathcal{P}_t=\{p_m\}_{m=1}^M$ where $p_m$ represents node position.

\subsection{Control}\label{sec:control}
The control module is responsible for converting the topological plan $\mathcal{P}_t$ into a series of low-level actions that guide the agent to the goal. 
Inputs for the control module include a sequence of subgoal nodes spanning $\mathcal{P}_t$, and the agent's pose at each each time step.
The output action space of navigation control is a set of parameterized low-level actions defined by the VLN-CE task, \eg, \textrm{FORWARD} (0.25m), \textrm{ROTATE LEFT/RIGHT} (15\degree), and \textrm{STOP}.

The control module produces actions that move the agent from one node to another.
Similar to~\cite{hong2022bridging,krantz2021waypoint}, we employ a rotate-then-forward controller, named RF.
Specifically, to reach a subgoal node $p_m$, RF first accesses the agent's current pose and computes its relative orientation and distance ($\triangle \theta, \triangle \rho$) from $p_m$. 
Then, ($\triangle \theta, \triangle \rho$) are quantized and translated to a series of \textrm{ROTATE} (15\degree) actions, followed by a \textrm{FORWARD} (0.25m) action sequence. 
RF executes these translated actions sequentially.
After that, the current subgoal is consumed and the subsequent node in plan $\mathcal{P}_t$ becomes the new subgoal.
The cycle repeats until no more nodes remain in $\mathcal{P}_t$.

\vspace{1mm}
\nbf{Handling Unreachable Goal.}
It is possible that the predicted long-term goal (a ghost node) is unreachable, due to its position being estimated by predicted waypoints that might not be on the navigation mesh. 
In such cases, there is a risk of the agent repeatedly selecting the same unreachable goal node in alternating planning stages, inevitably leading to no progress in navigation control. 
To alleviate this issue, we employ a simple strategy - delete the selected ghost node from the graph map $G_t$ before trying to reach it using navigation control. 
This approach not only avoids the repeated selection of unfeasible ghost nodes but also reduces the pool of candidates available for long-term goal prediction, thereby easing policy learning.

\vspace{1mm}
\nbf{Obstacle Avoidance.}
The VLN-CE task simulates a practical navigation scenario where collision with obstacles is taken into account.
Obstacle avoidance is essential, especially when sliding along obstacles is forbidden, such as on the RxR-CE dataset~\cite{ku2020room}.
In such cases, the agent is unable to move forward if its chassis comes into contact with an obstacle. 
This can result in navigation deadlocks, and in extreme cases, navigation failure due to early episode termination.
To address this issue, we devise a trial-and-error heuristic called `Tryout' to prevent navigation deadlocks. 
`Tryout' comes into play when the RF controller executes the \textrm{FORWARD} flow, and it shares a similar spirit with the `brute force untrap mode'~\cite{luo2022stubborn}.
%
Concretely, it detects deadlocks by checking if the agent's position changes after executing a \textrm{FORWARD} action. 
If a deadlock is identified, the Tryout compels the agent to rotate with a set of predefined headings $\triangle \Theta^\text{try}$ and attempt to move on with a single \textrm{FORWARD} action. 
If the agent moves away from its previous position, it indicates that the agent has exited the dead-end.
Then, the agent returns to its original heading and continues with the remaining \textrm{FORWARD} control flow.
However, if the agent's position remains unchanged, it proceeds to try other headings in $\triangle \Theta^\text{try}$.
In practice, $\triangle \Theta^\text{try}$ consists of 7 equally-space horizontal heading angles, ranging from 90\degree counterclockwise ($-90\degree$) to 90\degree clockwise ($90\degree$).

\subsection{Training and Inference}\label{sec:training}
\nbf{Pre-training.}
We pre-train the planning module with proxy tasks to improve its generalization ability, following the common practice in transformer-based VLN models~\cite{majumdar2020improving,hong2021vln,chen2021history}.
In this stage, the input topo maps are constructed offline and derived from the predefined graphs used in Matterport3D simulator~\cite{anderson2018vision}. 
Specifically, given an expert trajectory, we first extract its corresponding sub-graph from the predefined graph, and then mark the current node, visited nodes, and ghost nodes along the trajectory. 
Further, we align rendered RGBD images in the Habitat Simulator~\cite{savva2019habitat} onto the predefined graph for feature mapping in the map construction process. 
We adopt Masked Language Modeling (MLM)~\cite{kenton2019bert} and Single Action Prediction (SAP)~\cite{hao2020towards} proxy tasks for pre-training.
In the MLM task, the input instructions are randomly masked and the planning module is optimized by recovering the masked words after map-instruction interaction as described in \S~\ref{sec:planning}. 
As for the SAP task, we randomly chunk an input expert trajectory and build its corresponding topo map. The objective of this task is to predict the next teacher action, \ie, the subsequent action node of the chunked trajectory.

\vspace{1mm}
\nbf{Fine-tuning.}
We then fine-tune our model on downstream VLN-CE tasks to adapt navigation on 3D meshes in the Habitat Simulator~\cite{savva2019habitat}. 
To avoid overfitting to expert experience, we use `student-forcing'~\cite{krantz2020beyond} to train the model, where the predicted long-term goal of each step is sampled through the probability distribution of the predicted scores (Eq.~\ref{eq:predict}). 
In each decision loop, the agent updates the topo map as described in \S~\ref{sec:mapping}, and then conducts cross-modal map-instruction reasoning to predict a long-term goal as explained in \S~\ref{sec:planning}. 
The planned path is executed by a controller as presented in \S~\ref{sec:control}. 
Similar to DAgger~\cite{ross2011reduction}, we employ an interactive demonstrator $*$ to determine the teacher action node of each step. 
The demonstrator $*$ accesses the ground-truth 3D mesh and outputs one ghost node as the teacher supervision. 
$*$ takes different strategies on different datasets. 
On the R2R-CE dataset, the teacher action node $a_{t}^{*}$ is the ghost node that has the shortest geodesic distance to the final target. 
On the RxR-CE dataset that does not has shortest-path prior, $*$ takes a path-fidelity strategy similar to~\cite{raychaudhuri2021language}. Concretely, the annotated reference path is discretized as a sequence of subgoals. $*$ keeps track of visited subgoals and $a_{t}^{*}$ is the ghost node that has the shortest geodesic distance to the next unvisited subgoal. 
Overall, the policy learning objective is formulated as:
\vspace{-1mm}
\begin{equation}
L = \sum_{t=1}^T -\log p(a_{t}^{*} | W, G_t)
\end{equation}\label{eq:finetune}
\vspace{-2mm}

\nbf{Inference.}
During the testing phase, the agent consistently runs the mapping-planning-control cycle, which is analogous to the fine-tuning stage. 
The primary distinction between the two stages pertains to the long-term goal sampling strategy employed at each planning step.
In this case, the agent greedily selects the ghost node with the maximum predicted scores (Eq.~\ref{eq:predict}). 
In the event that the agent triggers a \textrm{STOP} action or surpasses the maximum number of goal predictions, the ongoing episode's navigation will terminate. 
In line with~\cite{hong2022bridging}, the maximum number of predictions is set as 15 for the R2R-CE dataset and 25 for the RxR-CE dataset.

\begin{table*}[htbp]
\renewcommand{\arraystretch}{1.2}
\centering
\caption{Data statistics and agent embodiment of R2R-CE and RxR-CE datasets.}
\vspace{-2mm}
\label{tab:data_analysis}%
\resizebox{\textwidth}{!}{
\begin{tabular}{lc|cc|cc|cc|cc|cc|cc}
\hline
\multirow{2}[2]{*}{Dataset} & \multirow{2}[2]{*}{Language} & \multicolumn{2}{c|}{Length} & \multicolumn{2}{c|}{Train} & \multicolumn{2}{c|}{Val-Seen} & \multicolumn{2}{c|}{Val-Unseen} & \multicolumn{2}{c|}{Test-Unseen} & \multicolumn{2}{c}{Embodiment } \\
\cline{3-4} \cline{5-6} \cline{7-8} \cline{9-10} \cline{11-12} \cline{13-14}  
&       & Path  & Sentence & \#house & \#instr & \#house & \#instr & \#house & \#instr & \#house & \#instr & Chassis & Sliding \\
\hline
R2R-CE & en         & 9.89m  & 32 words    & 61    & 10,819 & 53    & 778    & 11    & 1,839  & 18    & 3,408  & 0.10m  & Allowed  \\
RxR-CE & en, hi, te & 15.23m & 120 words   & 59    & 60,300 & 57    & 6,746  & 11    & 11,006 & 17    & 9,557  & 0.18m  & Forbidden  \\
\hline
\end{tabular}
}
\vspace{-2mm}
\end{table*}%
\section{Experiment}\label{sec:exprs}
\subsection{Experimental Setup}
\subsubsection{Datasets}
We conduct experiments on R2R-CE and RxR-CE datasets, which are created by converting discrete paths of R2R~\cite{anderson2018vision} and RxR~\cite{ku2020room} datasets into continuous environments through the Habitat Simulator~\cite{savva2019habitat}. 
While both datasets provide step-by-step language guidance, they differ in various aspects such as path length, guidance granularity, and agent embodiment as summarized in Tab.~\ref{tab:data_analysis}. 

The R2R-CE dataset comprises a total of 5,611 shortest-path trajectories, encompassing train, validation, and test splits.
Each trajectory corresponds to approximately 3 English instructions. 
The average path length is 9.89m and each instruction consists of an average of 32 words. 
We report performance on several validation splits. Val-Seen contains episodes with novel paths and instructions but from scenes observed in training. 
Val-Unseen contains novel paths, instructions, and scenes. 
Agents in R2R-CE have a chassis radius of 0.10m and can slide along obstacles while navigating.

RxR-CE is larger and more challenging compared to R2R-CE. 
While having similar scene splits as R2R-CE, RxR-CE presents substantively more instructions, spanning multilingual descriptions in English, Hindi, and Telugu, requiring an average of 120 words per instruction. 
Additionally, annotated paths in RxR-CE are much longer than those in R2R-CE (15.23m \textit{v.s.} 9.89m). 
To be noted, agents in RxR-CE are forbidden to slide along obstacles, and the larger chassis radius (0.18m) makes it prone to collide with obstacles. 
This also makes RxR-CE more challenging because navigation can easily get stuck when encountering obstacles, underscoring the vital role of obstacle avoidance in this challenging task.

\subsubsection{Evaluation Metrics}
Following previous works~\cite{anderson2018vision,anderson2018evaluation,ilharco2019general}, we adopt the following navigation metrics. Trajectory Length (TL): average path length in meters; Navigation Error (NE): average geometric distance in meters between the final and target location; Success Rate (SR): the ratio of paths with NE less than 3 meters; Oracle SR (OSR): SR given oracle stop policy; SR penalized by Path Length (SPL); Normalize Dynamic Time Wrapping (NDTW): the fidelity between the predicted and annotated paths and NDTW penalized by SR (SDTW).  R2R-CE uses SR and SPL as its primary metrics, whereas RxR-CE is more concerned with path fidelity and uses NDTW and SDTW as its primary metrics.

\subsubsection{Implementation Details}\label{sec:imp}
\nbf{Model Configuration.}
For visual encoding, we use ViT-B/32~\cite{dosovitskiy2020image} pre-trained in CLIP~\cite{radford2021learning} to encode RGB images as ~\cite{an20221st}, and ResNet-50~\cite{he2016deep} pre-trained in point-goal navigation~\cite{wijmans2019dd} to encode depth images following ~\cite{krantz2020beyond}.
The same as~\cite{chen2021history,tan2019lxmert,hong2022bridging}, we set the layers’ number of the panorama encoder, the text encoder, and the cross-modal graph encoder as 2, 9, 4, respectively.
Other hyperparameters are the same as LXMERT~\cite{tan2019lxmert} (\eg, the hidden layer size is 768). 
In the pre-training stage, we initialize the model with pre-trained LXMERT on the R2R-CE dataset and pre-trained RoBerta~\cite{liu2019roberta} for the multilingual RxR-CE dataset.

\begin{table*}[h]
\renewcommand{\arraystretch}{1.2}
\centering
\caption{Comparison with state-of-the-art methods on R2R-CE dataset.}\label{tab:sota_r2r}
\vspace{-2mm}
\resizebox{\textwidth}{!}{
\begin{tabular}{l | ccccc | ccccc | ccccc}
\hline
\multirow{2}[2]{*}{Methods} & \multicolumn{5}{c|}{Val Seen} & \multicolumn{5}{c|}{Val Unseen} & \multicolumn{5}{c}{Test Unseen} \\
\cline{2-6} \cline{7-11} \cline{12-16} & 
\multicolumn{1}{c}{TL} & \multicolumn{1}{c}{NE$\downarrow$} & \multicolumn{1}{c}{OSR$\uparrow$} & \multicolumn{1}{c}{\textbf{SR}$\uparrow$} & \multicolumn{1}{c|}{\textbf{SPL}$\uparrow$} &
\multicolumn{1}{c}{TL} & \multicolumn{1}{c}{NE$\downarrow$} & \multicolumn{1}{c}{OSR$\uparrow$} & \multicolumn{1}{c}{\textbf{SR}$\uparrow$} & \multicolumn{1}{c|}{\textbf{SPL}$\uparrow$} & 
\multicolumn{1}{c}{TL} & \multicolumn{1}{c}{NE$\downarrow$} & \multicolumn{1}{c}{OSR$\uparrow$} & \multicolumn{1}{c}{\textbf{SR}$\uparrow$} & \multicolumn{1}{c}{\textbf{SPL}$\uparrow$} \\
\hline
Seq2Seq~\cite{krantz2020beyond}
& 9.26 & 7.12 & 46 & 37 & 35
& 8.64 & 7.37 & 40 & 32 & 30
& 8.85 & 7.91 & 36 & 28 & 25  \\
SASRA~\cite{irshad2021sasra}
& 8.89 & 7.71 & - & 36 & 34 
& 7.89 & 8.32 & - & 24 & 22
& - & - & - & - & - \\
CMTP~\cite{chen2021topological}
& - & 7.10 & 56 & 36 & 31 
& - & 7.90 & 38 & 26 & 23
& - & - & - & - & - \\
LAW~\cite{raychaudhuri2021language}
& - & - & - & 40 & 37 
& - & - & - & 35 & 31
& - & - & - & - & - \\
HPN~\cite{krantz2021waypoint}
& 8.54 & 5.48 & 53 & 46 & 43 
& 7.62 & 6.31 & 40 & 36 & 34
& 8.02 & 6.65 & 37 & 32 & 30 \\
CM2~\cite{georgakis2022cross}
& 12.05 & 6.10 & 51 & 43 & 35 
& 11.54 & 7.02 & 42 & 34 & 28
& 13.90 & 7.70 & 39 & 31 & 24 \\
WS-MGMAP~\cite{chen2022weakly}
& 10.12 & 5.65 & 52 & 47 & 43 
& 10.00 & 6.28 & 48 & 39 & 34
& 12.30 & 7.11 & 45 & 35 & 28 \\
CWP-CMA~\cite{hong2022bridging}
& 11.47 & 5.20 & 61 & 51 & 45 
& 10.90 & 6.20 & 52 & 41 & 36
& 11.85 & 6.30 & 49 & 38 & 33 \\
CWP-RecBERT~\cite{hong2022bridging}
& 12.50 & 5.02 & 59 & 50 & 44 
& 12.23 & 5.74 & 53 & 44 & 39
& 13.51 & 5.89 & 51 & 42 & 36 \\
Sim2Sim~\cite{krantz2022sim}
& 11.18 & 4.67 & 61 & 52 & 44 
& 10.69 & 6.07 & 52 & 43 & 36
& 11.43 & 6.17 & 52 & 44 & 37 \\
\hline
Reborn (ours)~\cite{an20221st}
& 10.29 & 4.34 & 67 & 59 & 56 
& 10.06 & 5.40 & 57 & 50 & 46
& 11.47 & 5.55 & 57 & 49 & 45 \\
ETPNav (ours)
& 11.78 & 3.95 & 72 & \textbf{66} & \textbf{59} 
& 11.99 & 4.71 & 65 & \textbf{57} & \textbf{49} 
& 12.87 & 5.12 & 63 & \textbf{55} & \textbf{48}  \\
\hline
\end{tabular}
}
\end{table*}

\vspace{1mm}
\nbf{Training Details.}
Our experiments were performed using the PyTorch framework~\cite{paszke2019pytorch} and executed on two NVIDIA RTX 3090 GPUs. 
Our model includes two trainable modules: the panorama encoder used in the topological mapping module, and the cross-modal planning module.
We pre-train our model for 100,000 iterations ($\sim$ 20 hours), with a batch size of 64 and a learning rate of 5e-5, utilizing the AdamW optimizer~\cite{loshchilovdecoupled}. 
In this stage, topological maps are built offline and derived from the predefined graph of discrete VLN~\cite{anderson2018vision}. 
We leverage the discrete paths in the R2R and RxR datasets for pre-training purposes and augment the data using synthetic instructions from Prevalent~\cite{hao2020towards} and RxR-Markey~\cite{wang2022less}. 
After pre-training, we choose the model weights, producing the best zero-shot navigation performance (\eg, SPL on R2R-CE, SDTW on RxR-CE) to initialize the fine-tuning stage. 
During fine-tuning, the agent interacts with the environments online through the Habitat Simulator~\cite{savva2019habitat} and is supervised by the teacher node generated by the demonstrator $*$. 
We leverage scheduled sampling~\cite{bengio2015scheduled} to train the model, shifting from teacher-forcing to student-forcing with a decay frequency of per 3000 iterations and decay ratio of 0.75.
The fine-tuning iterations amount to 15,000 ($\sim$ 30 hours) with a batch size of 16 and a learning rate of 1e-5. 
The best iterations are determined by best performance on validation unseen splits.

\begin{table}[t]
\renewcommand{\arraystretch}{1.1}
\centering
\caption{Comparison against other VLN planners.}\label{tab:sota_planner}
\vspace{-2mm}
\setlength\tabcolsep{8pt}
\resizebox{0.45\textwidth}{!}{
\begin{tabular}{c | ccccc}
\hline
\multirow{2}[2]{*}{Planners} & \multicolumn{5}{c}{R2R-CE Val-Unseen} \\
\cline{2-6}
& TL & NE$\downarrow$ & OSR$\uparrow$ & \textbf{SR}$\uparrow$ & \textbf{SPL}$\uparrow$ \\
\hline
RecBERT~\cite{hong2022bridging} & 12.46 & 5.66  & 54.13 & 45.14 & 39.87 \\
HAMT~\cite{chen2021history}     & 10.23 & 5.41  & 57.09 & 49.97 & 45.72 \\
DUET~\cite{chen2022think}       & 13.08 & 5.16  & 62.21 & 53.71 & 46.43 \\
ETPNav (Ours)                   & 11.99 & 4.71  & 64.71 & \textbf{57.21} & \textbf{49.15} \\
\hline
\end{tabular}
}
\vspace{-2mm}
\end{table}

\begin{table*}[h]
\renewcommand{\arraystretch}{1.2}
\centering
\caption{Comparison with state-of-the-art methods on RxR-CE dataset.}\label{tab:sota_rxr}
\vspace{-2mm}
\resizebox{\textwidth}{!}{
\begin{tabular}{l | ccccc | ccccc | ccccc}
\hline 
\multirow{2}[2]{*}{Methods} & \multicolumn{5}{c|}{Val Seen} & \multicolumn{5}{c|}{Val Unseen} & \multicolumn{5}{c}{Test Unseen} \\
\cline{2-6} \cline{7-11} \cline{12-16} &
\multicolumn{1}{c}{NE$\downarrow$} & \multicolumn{1}{c}{SR$\uparrow$} & \multicolumn{1}{c}{SPL$\uparrow$} & \multicolumn{1}{c}{\textbf{NDTW}$\uparrow$} & \multicolumn{1}{c|}{\textbf{SDTW}$\uparrow$} &
\multicolumn{1}{c}{NE$\downarrow$} & \multicolumn{1}{c}{SR$\uparrow$} & \multicolumn{1}{c}{SPL$\uparrow$} & \multicolumn{1}{c}{\textbf{NDTW}$\uparrow$} & \multicolumn{1}{c|}{\textbf{SDTW}$\uparrow$} &
\multicolumn{1}{c}{NE$\downarrow$} & \multicolumn{1}{c}{SR$\uparrow$} & \multicolumn{1}{c}{SPL$\uparrow$} & \multicolumn{1}{c}{\textbf{NDTW}$\uparrow$} & \multicolumn{1}{c}{\textbf{SDTW}$\uparrow$} \\
\hline
Seq2Seq~\cite{krantz2020beyond}
& - & - & - & - & - 
& - & - & - & - & - 
& 12.10 & 13.93 & 11.96 & 30.86 & 11.01 \\
CWP-CMA~\cite{hong2022bridging}
& - & - & - & - & - 
& 8.76 & 26.59 & 22.16 & 47.05 & - 
& 10.40 & 24.08 & 19.07 & 37.39 & 18.65 \\
CWP-RecBERT~\cite{hong2022bridging}
& - & - & - & - & - 
& 8.98 & 27.08 & 22.65 & 46.71 & - 
& 10.40 & 24.85 & 19.61 & 37.30 & 19.05 \\
\hline
Reborn\dag (ours)~\cite{an20221st}
& 5.73 & 51.14 & 44.78 & 65.72 & 43.84
& 5.82 & 47.56 & 41.65 & 63.02 & 41.16
& - & - & - & - & - \\
Reborn (ours)~\cite{an20221st}
& 5.69 & 52.43 & 45.46 & 66.27 & 44.47 
& 5.98 & 48.60 & 42.05 & \textbf{63.35} & 41.82 
& 7.10 & 45.82 & 38.82 & \textbf{55.43} & 38.42 \\
ETPNav\dag (ours)
& 5.55 & 57.26 & 47.67 & 64.15 & 47.57 
& 5.80 & 53.07 & 44.16 & 61.49 & 43.92 
& - & - & - & - & - \\
ETPNav (ours)
& 5.03 & 61.46 & 50.83 & \textbf{66.41} & \textbf{51.28} 
& 5.64 & 54.79 & 44.89 & 61.90 & \textbf{45.33} 
& 6.99 & 51.21 & 39.86 & 54.11 & \textbf{41.30} \\
\hline
\multicolumn{16}{l}{\small{\dag Results without Marky-mT5 synthetic instructions~\cite{wang2022less}.}}
\end{tabular}
}
\vspace{-2mm}
\end{table*}
\subsection{Comparison with State-of-the-art Methods}
\subsubsection{R2R-CE}
In Tab.~\ref{tab:sota_r2r}, we compared our ETPNav with current state-of-the-art methods on the R2R-CE dataset.
The results demonstrate that our model outperforms the existing models on all splits in terms of NE, OSR, SR, and SPL.
Particularly, on the val unseen split, ETPNav surpasses the second-best model CWP-RecBERT~\cite{hong2022bridging} by 13 SR and 10 SPL. Moreover, our model also generalizes well on the test unseen split, as it outperforms Sim2Sim~\cite{krantz2022sim} by 11 SR and 11 SPL.
Reborn~\cite{an20221st} serves as the initial version for the 2022 RxR-Habitat Challenge. It uses a local planning space, which consists of nearby waypoints, and utilizes an unstructured memory bank to capture navigation dependency. 
The performance gap between Reborn and ETPNav is substantial, with ETPNav outperforming Reborn on the test unseen split by 6 SR and 3 SPL.
This highlights the efficacy of global planning with topo maps, enabling the agent to encode structured environmental priors and allowing for long-term planning, leading to a more robust policy.
We also compare ETPNav against several advanced VLN planners in Tab.~\ref{tab:sota_planner}. 
Specifically, RecBERT~\cite{hong2022bridging}, HAMT~\cite{chen2021history} and DUET~\cite{chen2022think} are transferred into VLN-CE where we align their controllers and model configurations with ETPNav's (\eg, the CLIP-ViT-B/32~\cite{radford2021learning} visual encoder, depth-only waypoint and topological map). 
ETPNav consistently surpasses these alternatives by a clear margin (\eg, 3.5 SR and 2.72 SPL higher than DUET). 
This further demonstrates the efficacy of ETPNav's planning capability.

\subsubsection{RxR-CE}
Tab.~\ref{tab:sota_rxr} compares our ETPNav model with the current state-of-the-art methods on the RxR-CE dataset.
Our model outperforms the existing best model, CWP-RecBERT~\cite{hong2022bridging} on all evaluation metrics on the three splits. For instance, on the val unseen split, ETPNav surpasses CWP-RecBERT by 27.71 SR, 22.24 SPL, and 15.19 NDTW.
ETPNav also generalizes well on the test unseen split, where it outperforms CWP-RecBERT by 26.36 SR, 20.25 SPL, 16.81 NDTW, and 22.25 SDTW. 
For a fair comparison, we also report our results without Marky-mT5~\cite{wang2022less} data augmentation, where ETPNav still beats CWP-RecBERT by a significant margin, for instance, $\uparrow$25.99 SR and $\uparrow$14.78 NDTW on the val unseen split.
Please note that Reborn~\cite{an20221st} is our winning entry for the 2022 RxR-Habitat Challenge, which employs a local planning space composed of nearby waypoints. While Reborn achieves slightly better NDTW (\eg, 55.43 \textit{v.s.} 54.11) on the test unseen splits, it has significant worse SDTW (\eg, 38.43 \textit{v.s.} 41.30).
We attribute this to the global planning space of ETPNav, which promotes backtracking and may impact path fidelity. However, this global planning space enables the agent to make long-term plans, resulting in better SR and SDTW.

\subsection{Ablation Study}
In this section, we provide detailed ablation experiments to evaluate specific components of ETPNav, including critical design choices of the topological mapping module (\S~\ref{sec:ab_map}) and the cross-modal planning module (\S~\ref{sec:ab_plan}). 
Additionally, we compare the proposed controller with other alternatives (\S~\ref{sec:ab_control}). 
Finally, we visualize the trajectories predicted by our model and compare them with other variants (\S~\ref{sec:qualitative}).

\begin{table*}[htbp]
\renewcommand{\arraystretch}{1.2}
\centering
\caption{Comparison of different waypoint predictors.}\label{tab:ab_predictor}
\vspace{-2mm}
\resizebox{0.92\textwidth}{!}{
\begin{tabular}{cc | cccc | ccccc | ccccc}
\hline
\multirow{2}[2]{*}{\#} & \multirow{2}[2]{*}{inputs} & \multicolumn{4}{c|}{Waypoint Prediction} & \multicolumn{5}{c|}{R2R-CE Val-Seen} & \multicolumn{5}{c}{R2R-CE Val-Unseen} \\
\cline{3-6} \cline{7-11} \cline{12-16} 
&  & $|\triangle|$ & \%Open$\uparrow$  & $d_C \downarrow$  & $d_H \downarrow$  & TL    & NE$\downarrow$    & OSR$\uparrow$   & \textbf{SR}$\uparrow$    & \textbf{SPL}$\uparrow$   & TL    & NE$\downarrow$    & OSR$\uparrow$   & \textbf{SR}$\uparrow$    & \textbf{SPL}$\uparrow$ \\
\hline
1 & RGBD  & 1.40  & 82.87 & 1.05  & \textbf{2.01}  & 11.22 & 3.87  & 70.56 & 63.88 & 57.11 & 11.77 & 4.73  & 63.24 & 56.44 & 48.53 \\
2 & RGB   & 1.38  & 65.34 & 1.08  & 2.03  & 13.38 & 4.38  & 63.49 & 56.29 & 46.42 & 12.81 & 4.99  & 57.91 & 51.66 & 42.21 \\
3 & Depth & 1.39  & \textbf{84.05} & \textbf{1.04}  & \textbf{2.01}  & 11.78 & 3.95  & 71.85 & \textbf{66.19} & \textbf{59.37} & 11.99 & 4.71  & 64.71 & \textbf{57.21} & \textbf{49.15} \\
\hline
\end{tabular}
}
\vspace{-2mm}
\end{table*}%
\subsubsection{Key Design Choices of Topological Mapping}\label{sec:ab_map}
\nbf{Waypoint Prediction.}
Tab.~\ref{tab:ab_predictor} presents a comparison between three different waypoint predictors on the R2R-CE dataset.
In Row 1, RGB and depth features are utilized as inputs, where both feature types are linearly transformed to the same dimension, fused, and then fed into the transformer layer to predict waypoints. This approach is also the default choice in~\cite{hong2022bridging}.
Row 2 only takes RGB features as inputs, while Row 3 shows our approach that uses only depth features for waypoint prediction. 
We apply waypoint metrics~\cite{hong2022bridging} and navigation results to assess the quality of predicted waypoints.
These waypoint metrics are as follows: $|\triangle|$ measures the difference in the number of target waypoints and predicted waypoints. \%Open measures the ratio of waypoints that is in open space (not hindered by any obstacle). $d_C$ and $d_H$ are the Chamfer distance and the Hausdorff distance, respectively, commonly used metrics to measure the distance between point clouds. 

As shown in Tab.~\ref{tab:ab_predictor}, Row 1 achieves a decent performance in both waypoints and navigation metrics on the val unseen split, with 82.87 \%Open, 1.05 $D_C$, and 56.44 SR. 
Conversely, Row 2 only utilizes RGB to predict waypoints, resulting in the worst performance of all with 65.34 \%Open and 1.08 $D_C$. 
Without depth information, the \%Open metric drops severely, indicating that many waypoints are obstructed by obstacles or not on the navigation mesh.
Consequently, the navigation performance also declines considerably, for example, compared to Row 1, SR drops by 4.78 on the val unseen split. 
It is noteworthy that the depth-only predictor (Row 3) yields the best performance, achieving 84.05 \%Open and 1.04 $D_C$. 
Additionally, the navigation performance is also superior, with 57.21 SR  and 49.15 SPL on the val unseen split, compared to 56.44 SR and 48.53 SPL respectively. 
These findings suggest that RGB information is ineffective and even detrimental to waypoint prediction. 
One possible explanation is that low-level semantics in RGB features can make the predictor overfit to seen environments, while such semantics are unnecessary for inferring spatial accessibility.

\begin{table}[t]
\renewcommand{\arraystretch}{1.2}
\centering
\caption{Comparison of different options for map construction. 
$\gamma$ is the threshold of the waypoint localization function $\mathcal{F}_L$.
`Acc.' denotes accumulating multiple waypoints to represent one ghost node.
`Del.' denotes deleting the selected ghost node in each planning step. 
$N_{\textrm{node}}$ denotes the average number of nodes per episode.  }\label{tab:ab_map}
\vspace{-2mm}
\resizebox{0.49\textwidth}{!}{\begin{tabular}{cccc | cccccc}
\hline
\multirow{2}[2]{*}{\#} & \multirow{2}[2]{*}{$\gamma$ (m)} & \multirow{2}[2]{*}{Acc.} & \multirow{2}[2]{*}{Del.} & \multicolumn{6}{c}{R2R-CE Val-Unseen} \\
\cline{5-10}
& & & & $N_{\textrm{node}}$ & TL & NE$\downarrow$ & OSR$\uparrow$ & \textbf{SR}$\uparrow$ & \textbf{SPL}$\uparrow$ \\
\hline
1 & \multirow{3}[2]{*}{0.25} & \checkmark      & \checkmark      & 32.18 & 11.68 & 4.70  & 63.34 & 56.71 & 48.71 \\
2 & & & \checkmark & 34.13 & 11.33 & 4.81  & 62.26 & 55.19 & 48.30 \\
3 & & \checkmark & & 30.80 & 13.46 & 5.08  & 62.15 & 53.34 & 45.64 \\
\hline
4 & \multirow{3}[2]{*}{0.50} & \checkmark & \checkmark & 23.76 & 11.99 & 4.71  & 64.71 & \textbf{57.21} & \textbf{49.15} \\
5 & & & \checkmark & 31.02 & 12.61 & 4.68  & 63.13 & 55.89 & 47.92 \\
6 & & \checkmark & & 21.01 & 13.38 & 5.03  & 59.70 & 52.41 & 45.02 \\
\hline
7 & \multirow{3}[2]{*}{0.75} & \checkmark & \checkmark & 18.23 & 13.97 & 4.94  & 64.11 & 54.75 & 45.42 \\
8 & & & \checkmark & 25.45 & 14.48 & 4.81  & 61.06 & 53.61 & 45.86 \\
9 & & \checkmark & & 15.57 & 15.04 & 5.07  & 58.78 & 51.16 & 42.38 \\
\hline
10 & \multirow{3}[2]{*}{1.00} & \checkmark & \checkmark & 14.43 & 18.60 & 6.13  & 52.31 & 42.30 & 33.60 \\
11 & & & \checkmark & 20.29 & 21.02 & 5.68  & 53.12 & 41.59 & 32.17 \\
12 & & \checkmark & & 10.98 & 18.52 & 5.50  & 49.32 & 42.36 & 33.09 \\
\hline
\end{tabular}}
\end{table}%
\vspace{1mm}
\nbf{Different Options for Map Construction.}
Tab.~\ref{tab:ab_map} compares different options for map construction on the R2R-CE dataset, including the localization threshold $\gamma$ and the waypoint accumulation in \S~\ref{sec:mapping}, as well as the ghost node deleting in \S~\ref{sec:control}.

As the localization threshold $\gamma$ increases, the number of nodes $N_{\textrm{node}}$ shows a downward trend. 
This is because a higher $\gamma$ encourages the agent to localize predicted waypoints onto existing nodes of the graph, thereby reducing the creation of new nodes. 
Meanwhile, the overall navigation performance is sensitive to the number of nodes $N_{\textrm{node}}$. For example, on the val unseen split, there is approximately 12 SR drop comparing (Row 10 $\sim$ Row 12) to (Row 1 $\sim$ Row 3). 
The reason is that a higher $\gamma$ results in too few nodes to depict the environment well, limiting the agent's accurate perception and efficient planning. 
However, a large $N_{\textrm{node}}$ also affects the navigation performance, \eg, on the val unseen split, 56.71 SR of Row 1 \textit{v.s.} 57.21 SR of Row 4. 
One potential reason is that a larger number of candidate nodes increases the learning difficulty of the planning module.

Moreover, both the `Acc.' and `Del.' are beneficial.
For instance, comparing Row 4 and Row 5, without `Acc.', SR and SPL on the val unseen split decrease by 1.32 and 1.23 respectively. 
`Acc.' allows the agent to integrate multi-step waypoint observations to represent ghost nodes, promoting the planning module to predict an accurate long-term goal.
Similarly, comparing Row 4 and Row 6, without `Del.', the performance decreases significantly, with SR and SPL on the val unseen split dropping by 4.80 and 4.13 respectively. 
Without `Del.', unreachable ghost nodes can be selected endlessly by the agent, resulting in no progress in navigation. 
In subsequent experiments, Row 4 is taken as the default setup.

\begin{table}[t]
\renewcommand{\arraystretch}{1.20}
\centering
\caption{Comparison of different plannig spaces (Local \textit{v.s.} Global). GASA represents graph-aware self-attention.}\label{tab:ab_act}
\vspace{-2mm}
\setlength\tabcolsep{6pt}
\resizebox{0.49\textwidth}{!}{
\begin{tabular}{ccc | ccccc}
\hline
\multirow{2}[2]{*}{\#} & \multirow{2}[2]{*}{\shortstack{Planning\\Space}} & \multirow{2}[2]{*}{GASA} & \multicolumn{5}{c}{R2R-CE Val-Unseen} \\
\cline{4-8}
& & & TL & NE$\downarrow$ & OSR$\uparrow$ & \textbf{SR}$\uparrow$ & \textbf{SPL}$\uparrow$ \\
\hline
1 & \multirow{2}[2]{*}{Local} & & 11.37 & 4.92  & 61.28 & 53.15 & 46.83 \\
2 & & \checkmark & 12.12 & 4.94  & 62.18 & 53.92 & 46.43 \\
\hline
3 & \multirow{2}[2]{*}{Global} & & 12.04 & 4.83  & 63.48 & 55.97 & 48.08 \\
4 & & \checkmark & 11.99 & 4.71  & 64.71 & \textbf{57.21} & \textbf{49.15} \\
\hline
\end{tabular}
}
\end{table}%
\subsubsection{Key Design Choices of Cross-Modal Planning}\label{sec:ab_plan}
\nbf{Comparison of Different Planning Space.}
Tab.~\ref{tab:ab_act} compares different planning spaces on the R2R-CE dataset as well as the effect of GASA in Eq.~\ref{eq:gasa}.
The local planning space only considers adjacent ghost nodes of the agent as candidate goals, while the global planning space consists of all observed ghost nodes along the traversed path. 
Global planning results in better navigation performance, \eg, on the val unseen split, Row 4 achieves a 57.21 SR compared to Raw 2 at 53.92 SR.
This demonstrates the superiority of global planning, as it allows efficient backtracking to a previous location, providing self-correction policy.
In contrast, local planning requires multiple plan-control flows to reach a remote location, introducing unstable accumulation bias, making it challenging to achieve such intelligent behavior.
GASA is also shown to be effective as it increases SR by about 1 point compared (Row 2, Row 4) to (Row 1, Row 3) where it is not used. 
GASA introduces topology for node encoding, facilitating the agent's ability to capture environmental structural priors.
We also note that the gain of GASA to global planning is more significant than that to local planning, comparing $\uparrow$ 1.24 SR in global planning and $\uparrow$ 0.77 SR in local planning.
We suspect this because conducting global planning requires an understanding of the house structure, while local planning is restricted to nearby areas, thereby reducing the need for structure priors.

\begin{table}[t]
\renewcommand{\arraystretch}{1.2}
\centering
\caption{The effect of pre-training tasks.}\label{tab:ab_pretrain_task}
\vspace{-2mm}
\setlength\tabcolsep{8pt}
\resizebox{0.49\textwidth}{!}{
\begin{tabular}{cc | ccccc}
\hline
\multirow{2}[2]{*}{\# } & \multirow{2}[2]{*}{Proxy Tasks} & \multicolumn{5}{c}{R2R-CE Val-Unseen} \\
\cline{3-7}
& & TL & NE$\downarrow$ & OSR$\uparrow$ & \textbf{SR}$\uparrow$ & \textbf{SPL}$\uparrow$ \\
\hline
1 & No init   & 14.35 & 6.81  & 49.21 & 37.41 & 30.28 \\
2 & MLM       & 16.69 & 5.44  & 58.51 & 48.23 & 37.73 \\
3 & SAP  & 13.05 & 5.04  & 59.98 & 52.37 & 42.77 \\
4 & MLM + SAP & 11.99 & 4.71  & 64.71 & \textbf{57.21} & \textbf{49.15} \\
\hline
\end{tabular}
}
\end{table}%
\nbf{The Effect of Pre-training.}
Tab.~\ref{tab:ab_pretrain_task} presents the benefits of various pre-training tasks on the downstream R2R-CE task.
Row 1 shows results from the model trained from scratch, which shows the worst performance (\eg, on the val unseen split, 37.41 SR and 30.28 SPL).
Row 2 displays the outcomes of pre-training with the MLM task, indicating a significant gain from the generic pre-training task (\eg, $\uparrow$ 10.82 SR and $\uparrow$ 7.45 SPL on the val unseen split). 
Because the MLM task enables the model to learn transferable visiolinguistic representations, enhancing the agent's generalization ability. 
Row 3 shows the efficacy of the SAP task. Compared to row 1, the navigation performance increases significantly with $\uparrow$14.96 SR and $\uparrow$12.49 SPL. 
Because SAP can promote the model to learn navigation-oriented representations, which is crucial for navigation policy learning. 
In row 4, the agent achieves the best performance when combining MLM and SAP tasks, demonstrating the two proxy tasks are complementary. 
Thus, we assemble MLM and SAP as our default pre-training tasks.  

\begin{table}[t]
\renewcommand{\arraystretch}{1.2}
\centering
\caption{Comparison of different visual inputs in pre-training.}\label{tab:ab_pretrain_input}
\vspace{-2mm}
\setlength\tabcolsep{8pt}
\resizebox{0.48\textwidth}{!}{
\begin{tabular}{ccc | ccccc}
\hline
\multirow{2}[2]{*}{\# } & \multirow{2}[2]{*}{RGB} & \multirow{2}[2]{*}{Depth} & \multicolumn{5}{c}{R2R-CE Val-Unseen} \\
\cline{4-8}
& & & TL & NE$\downarrow$ & OSR$\uparrow$ & \textbf{SR}$\uparrow$ & \textbf{SPL}$\uparrow$ \\
\hline
1 & MP3D  &              & 13.54 & 5.04  & 62.04 & 53.24 & 44.09 \\
2 & MP3D  & \checkmark   & 13.13 & 4.98  & 64.49 & 55.74 & 47.48 \\
\hline
3 & Habitat &            & 12.90 & 4.98  & 62.59 & 55.92 & 46.97 \\
4 & Habitat & \checkmark & 11.99 & 4.71  & 64.71 & \textbf{57.21} & \textbf{49.15} \\
\hline
\end{tabular}
}
\end{table}%
Tab.~\ref{tab:ab_pretrain_input} presents the effect of using different visual inputs for pre-training on the downstream R2R-CE task.
Row 1 and Row 2 pre-train the model with RGB images captured in Matterport3D Simulator~\cite{anderson2018vision}. which is a common practice in existing pre-training-based VLN-CE models~\cite{hong2022bridging,krantz2022sim,an20221st}.
In contrast, Row 3 and Row 4 use RGB images reconstructed in Habitat Simulator~\cite{savva2019habitat}. 
Notably, Row 3 outperforms Row 1 by 2.68 SR and 2.88 SPL on the val unseen split, highlighting a performance gap between MP3D and Habitat simulators due to their visual domain gap. 
Although the model can be fine-tuned using habitat images, the performance loss caused by the domain gap in the pre-training stage cannot be eliminated.
Additionally, adding depth information for pre-training (\eg, Row 4) shows better performance than Row 3, with an improvement of 1.29 SR and 2.18 SPL on the val unseen split. 
Therefore, Row 4 is used as our default pre-training setup.

\begin{table*}[htbp]
\renewcommand{\arraystretch}{1.2}
\centering
\caption{Comparison of different controllers.}\label{tab:ab_control}
\vspace{-2mm}
\setlength\tabcolsep{12pt}
\resizebox{0.95\textwidth}{!}{
\begin{tabular}{cc | ccccc | ccccc}
\hline
\multirow{2}[2]{*}{\#} & \multirow{2}[2]{*}{Controllers} & \multicolumn{5}{c|}{R2R-CE Val-Unseen} & \multicolumn{5}{c}{RxR-CE Val-Unseen} \\
\cline{3-7} \cline{8-12}
& & TL & NE$\downarrow$ & OSR$\uparrow$ & \textbf{SR}$\uparrow$ & \textbf{SPL}$\uparrow$ & NE$\downarrow$ & SR$\uparrow$ & SPL$\uparrow$ & \textbf{NDTW}$\uparrow$ & \textbf{SDTW}$\uparrow$ \\
\hline
1 & Teleportation & 11.31 & 4.64  & 65.14 & 57.97 & 49.76 & 5.80  & 54.98 & 45.23 & 64.33 & 46.04 \\
\hline
2 & PointGoal~\cite{wijmans2019dd} & 13.35 & 4.87 & 63.40 & 54.86 & 44.05 & 5.89 & 52.77 &  43.50 & 61.03 & 43.79 \\
3 & FMM~\cite{sethian1996fast} & 12.62 & 5.19 & 61.45 & 52.64 & 44.34 & 8.83 & 25.95 & 23.02 & 47.29 & 21.73 \\
4 & FMM \textit{w/} CMap  & 12.93 & 5.12	& 62.43	& 53.62	& 44.96 & 6.02 & 50.06 & 40.07 & 58.75 & 41.09 \\
5 & RF \textit{w/o} Tryout  & 11.99 & 4.71  & 64.71 & \textbf{57.21} & \textbf{49.15} & 9.22  & 22.61 & 20.06 & 44.30 & 18.64 \\
6 & RF \textit{w/} Tryout & -     & -     & -     & -     & -     & 5.64  & 54.79 & 44.89 & \textbf{61.90} & \textbf{45.33}  \\
\hline
\end{tabular}
}
\end{table*}%
\subsubsection{Comparison of Different Controllers}\label{sec:ab_control}
Tab.~\ref{tab:ab_control} compares the performance of various navigation controllers on the R2R-CE and RxR-CE datasets.
The Teleportation controller serves as the performance upper bound. It transports the agent to the goal predicted by the planning module.
Since the goal (a ghost node) might not be on the navigation mesh, in practice, we first transport the agent to an adjacent node of the goal, then drive it towards the goal using the proposed controller. 
Other alternative controllers include PointGoal, Fast Marching Method and RF.
PointGoal~\cite{wijmans2019dd} employs a pre-trained LSTM to predict low-level actions to reach the goal localization. 
We follow~\cite{georgakis2022cross} to adapt PointGoal to the VLN-CE task.
Fast Marching Method (FMM)~\cite{sethian1996fast} is a classical obstacle-aware controller which constructs a local occupancy map on the fly for path planning. This map is aggregated by the geometric projection of depth observations. 
We follow~\cite{krantz2022sim} to configure FMM on the VLN-CE task. 
We further consider an FMM variant (\ie, FMM \textit{w/} CMap) to handle invisible obstacles due to simulation error, where obstacle collisions are recorded as a collision map (CMap) similar to~\cite{luo2022stubborn}.
The CMap is then combined with the local occupancy map for path planning.
RF is the proposed controller described in \S~\ref{sec:control} and Tryout is active when sliding along obstacles is forbidden (\ie, RxR-CE dataset).

Row 1 establishes the upper bound, reaching 57.97 SR and 49.76 SPL on the R2R-CE dataset and 64.43 NDTW and 46.04 SDTW on the RxR-CE dataset.
In Row 2, PointGoal shows satisfactory performance, but there is a clear gap to Row 1 with a decrease of 5.71 SPL on the R2R-CE dataset and 2.25 SDTW on the RxR-CE dataset.
Row 3 and Row 4 show FMM could achieve decent performance on the R2R-CE dataset, whereas the collision map (Row 4) yields marginally superior outcomes. 
Moreover, the collision map shows to be critical on the sliding-forbidden RxR-CE dataset, where the performance drops severely without it (\eg, Row 3's 21.73 SDTW \textit{v.s.} Row 4's 41.09). 
Because the occupancy map alone is not enough for obstacle-avoiding, the agent in Row 3 gets stuck frequently. More analyses will be presented in the next paragraph. 
Row 5 shows that our RF controller manages to narrow the gap on the R2R-CE dataset, reaching 49.15 SPL compared to Row 1's 49.76 SPL. 
However, RF results in significant performance drops on the RxR-CE dataset with a 27.4 SDTW decrease compared to Row 1.
Because it is unaware of collision and causes frequent deadlocks in obstacles under the challenging sliding-forbidden setup, resulting in navigation failure. 
The proposed Tryout satisfactorily handles this problem, nearly eliminating the performance loss caused by sliding-forbidden with 45.33 SDTW on Row 6 compared to 46.04 SDTW on Row 1. 
Tryout even surpasses the learning-based PointGoal controller with 45.33 SDTW on Row 6 compared to 43.79 SDTW on Row 2.

\begin{table*}[htbp]
\renewcommand{\arraystretch}{1.2}
\centering
\caption{Quantitative results of the oracle planner with different controllers.}\label{tab:ab_control_detail}
\vspace{-2mm}
\resizebox{\textwidth}{!}{
\begin{tabular}{cc | ccccccc | ccccccc}
\hline
\multirow{2}[2]{*}{\#} & \multirow{2}[2]{*}{Controllers} & \multicolumn{7}{c|}{R2R-CE Val-Unseen} & \multicolumn{7}{c}{RxR-CE Val-Unseen} \\
\cline{3-9} \cline{10-16}
& & TL & AT & RT & \textbf{SG-NE}$\downarrow$ & \textbf{CT}$\downarrow$ & \textbf{SR}$\uparrow$ & \textbf{SPL}$\uparrow$ 
  & TL & AT & RT & \textbf{SG-NE}$\downarrow$ & \textbf{CT}$\downarrow$ & \textbf{SR}$\uparrow$ & \textbf{NDTW}$\uparrow$ \\
\hline
1 & PointGoal~\cite{wijmans2019dd} & 10.58 & 146.81 & 33.99 & 0.17  & 6.62  & 94.77 & 91.14 & 16.58 & 223.16 & 69.12 & 0.31  & 45.23 & 83.37 & 79.32 \\
2 & FMM~\cite{sethian1996fast} & 10.13 & 188.79 & 52.12 & 0.16  & 6.37  & 94.34 & 87.68 & 7.99 *  & 385.23 & 46.23 & 0.50  & 231.62 & 41.40 & 59.55 \\
3 & FMM \textit{w/} CMap & 10.20 & 188.44 & 52.64 & \textbf{0.15}  & 4.59  & 94.45 & 87.68 & 16.75 & 281.50 & 69.34 & 0.33  & 40.36 & 77.57 & 77.09 \\
4 & RF \textit{w/o} Tryout  & 9.56  & 60.63 & 21.18 & \textbf{0.15}  & \textbf{1.78}  & \textbf{97.61} & \textbf{93.27} & 7.49 * & 148.59 & 49.79 & 0.83  & 70.04 & 35.92 & 54.61 \\
5 & RF \textit{w/} Tryout & - & - & - & - & - & - & - & 15.46 & 148.58 & 64.80 & \textbf{0.28}  & \textbf{17.22} & \textbf{88.85} & \textbf{82.34} \\
\hline
\multicolumn{16}{l}{\small{* Much lower TL due to frequent collisions and deadlocks.}}
\end{tabular}
}
\end{table*}%
\begin{figure}[t]
\centering
\includegraphics[width=0.4\textwidth]{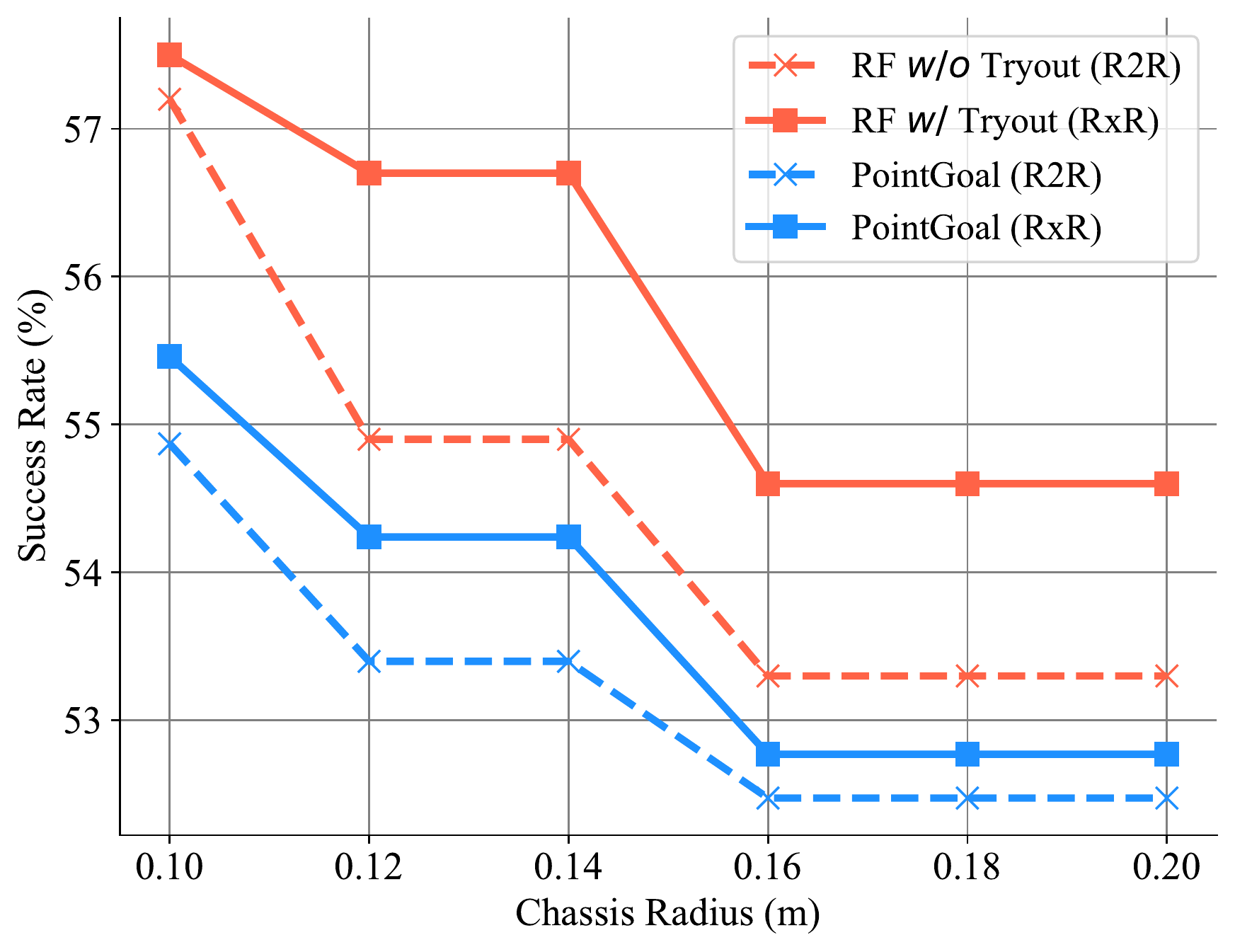}
\vspace{-2mm}
\caption{The effect of the agent's chassis radius on SR.}\label{fig:ab_radius}
\vspace{-3mm}
\end{figure}
In Tab.~\ref{tab:ab_control_detail}, we evaluate different controllers with an oracle planner, which always selects the teacher action node at each planning step.
Four extra metrics are considered, namely Actions Taken (AT), Rotations Taken (RT), Subgoal Navigation Error (SG-NE), and Collision Times (CT). 
AT and RT are the average numbers of actions and rotations (\textrm{ROTATE LEFT/RIGHT}) taken per episode.
SG-NE is the average distance between the teacher action node and the agent's position after control execution. 
CT denotes the average times of obstacle collisions per episode.
We summarize our findings below. 
First, compared to other controllers, RF variants (Row 4 and Row 5) take fewer actions and have shorter paths, which can contribute to their higher navigation efficiency (SPL) and path fidelity (NDTW). 
This is because the rotate-then-forward control flow forces the agent to walk in a straight line, while others may take unnecessary rotations to keep the agent away from obstacles. 
Second, though RF controllers simply walk in a straight line, they do not lead to significantly more collisions. 
We attribute this to the dataset~\cite{hong2022bridging} used to train our waypoint predictor, which makes the waypoints typically lie in straight-line accessible open spaces. This can also explain why Tryout can avoid deadlocks with a few attempts (Row 4 \textit{v.s.} Row 5).
Third, learning-based PointGoal can avoid obstacles implicitly (Row 1), while classical FMM failed to avoid obstacles and resulting in the highest CT (Row 2's 231.62 CT on the RxR-CE dataset). 
This indicates obstacle-avoiding relying on the occupancy map alone is difficult and a similar phenomenon is also observed in~\cite{luo2022stubborn}.
This is due to the imperfect occupancy map caused by obstacles out of view, inaccurate depth observations, and artifacts in scans of indoor spaces~\cite{krantz2020beyond}.
Row 3 mitigates this issue with collision information, and our Tryout also has a similar spirit.

\begin{figure*}[!htbp]
\centering
\includegraphics[width=0.92\textwidth]{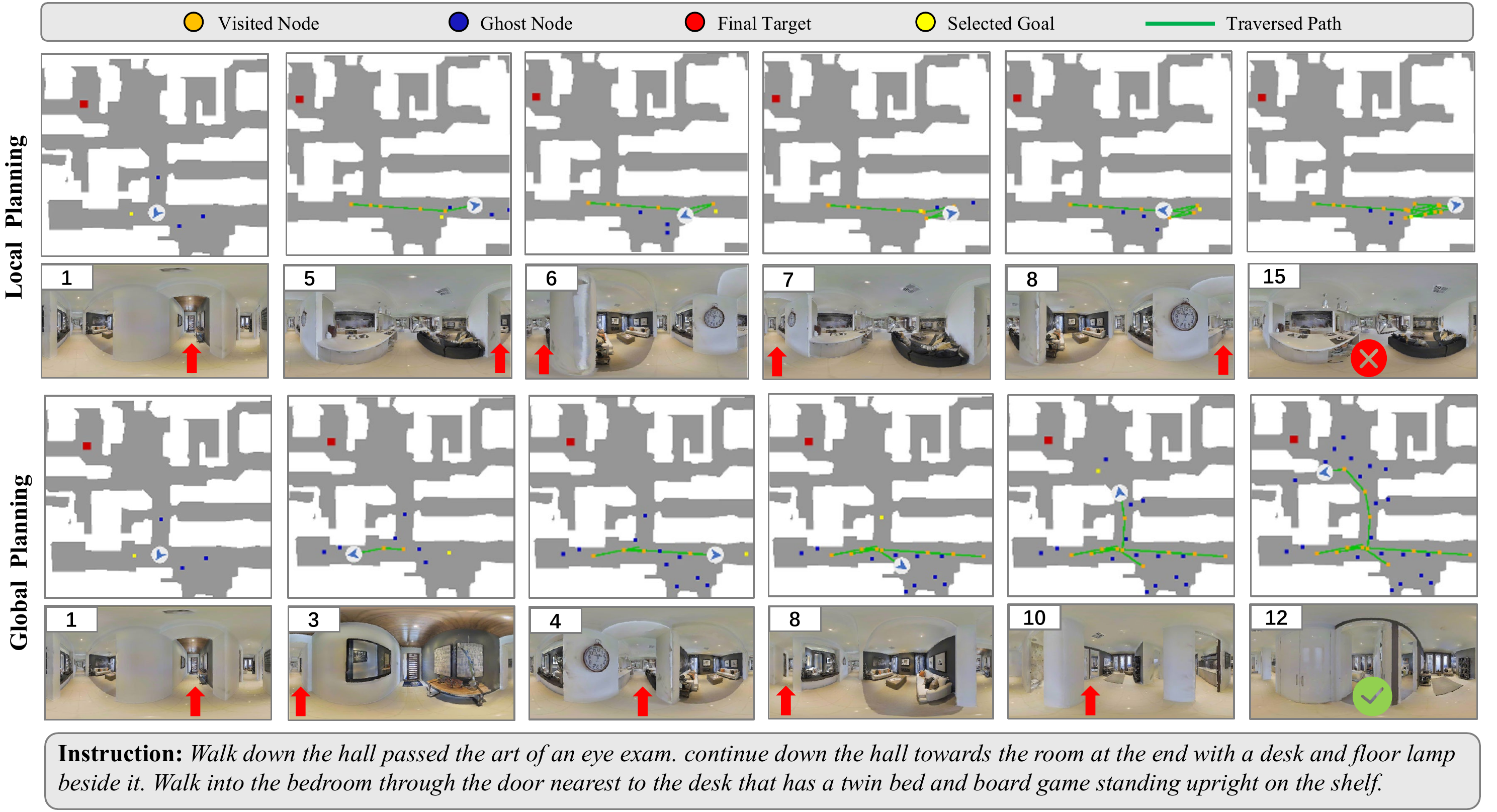}
\vspace{-2mm}
\caption{
Comparison of the same episode's trajectories predicted by different model variants. 
(Top) The trajectory predicted by ETPNav using local planning. 
(Bottom) The trajectory predicted by ETPNav using global planning. 
}\label{fig:vis1}
\vspace{-4mm}
\end{figure*}

\begin{figure*}[!htbp]
\centering
\includegraphics[width=0.92\textwidth]{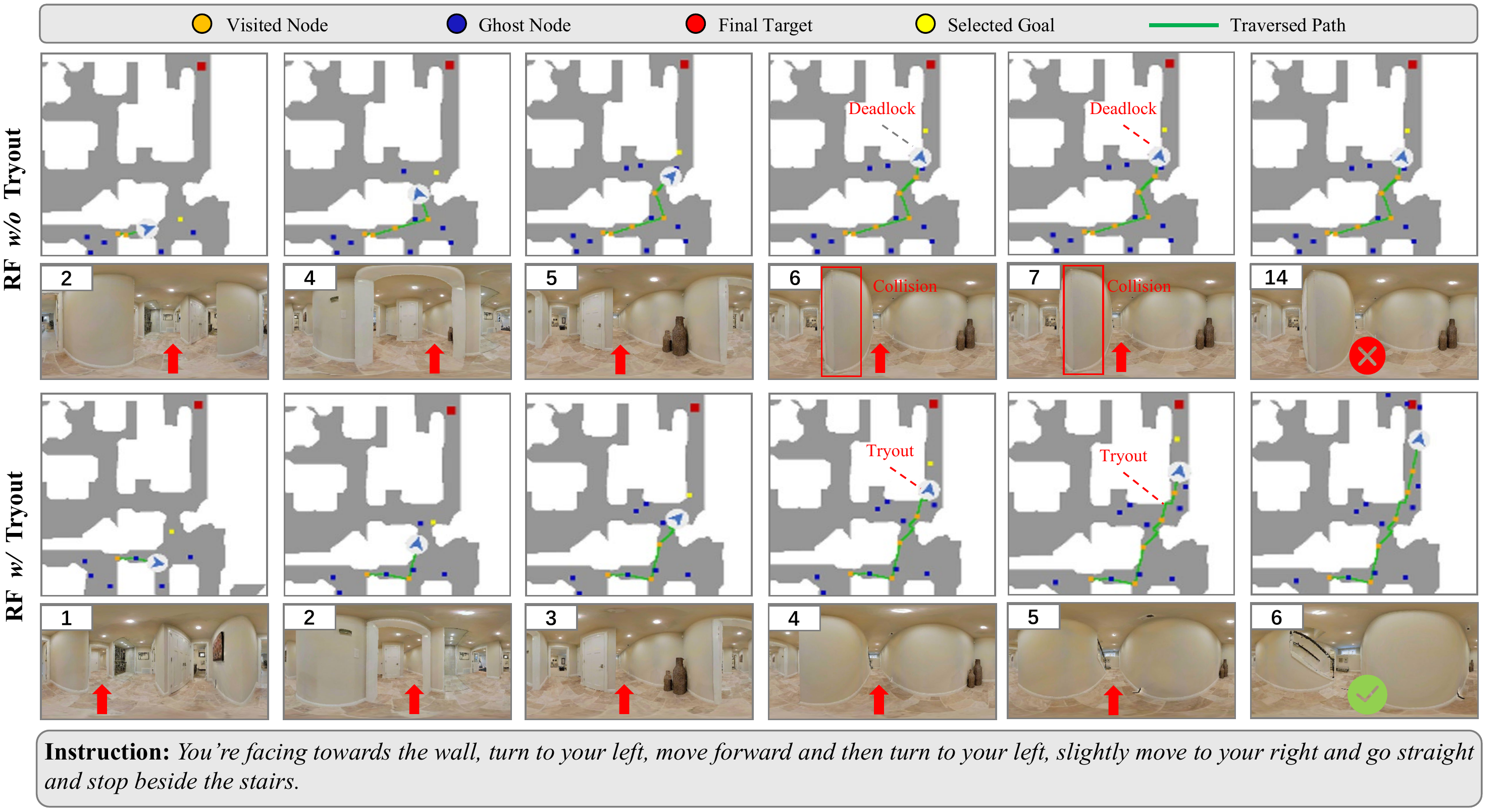}
\vspace{-2mm}
\caption{
Comparison of the same episode's trajectories predicted by different model variants. 
(Top) The trajectory predicted by ETPNav without Tryout control. 
(Bottom) The trajectory predicted by ETPNav with Tryout control. 
}\label{fig:vis2}
\vspace{-4mm}
\end{figure*}

Also, we are interested in how the agent's chassis radius may impact navigational performance.
Fig.~\ref{fig:ab_radius} shows the episodic success rate of ETPNav on the R2R- CE and RxR-CE datasets whether employing PointGoal or RF controllers. 
For all datasets, the success rate of the two controllers decreases as the chassis radius increases since a bigger chassis might lead the agent to collide with obstacles more frequently, increasing the risk of navigation failure. 
The proposed RF controller, however, is resilient to this problem and consistently outperforms the PointGoal controller, particularly on the RxR-CE dataset, where it beats PointGoal by about 3 SR through all chassis radiuses. 
The main reason is that it uses Tryout to explicitly prevent the agent from getting stuck in obstacles, which aids the adaptation of navigation policy to various chassis sizes. This further verifies the robustness of the proposed obstacle-avoiding controller.

\subsubsection{Qualitative Results}\label{sec:qualitative}
Fig.~\ref{fig:vis1} and Fig.~\ref{fig:vis2} visualize trajectories predicted by our model compared to the variant using local planning (on the R2R-CE dataset) and the variant without Tryout control (on the RxR-CE dataset), respectively. 

As shown in Fig.~\ref{fig:vis1}, the local planning space is insufficient to capture the global environment layouts and hinders the agent's long-term planning capacity. 
For example, at step 7, the agent seems to realize that it is navigating in the wrong direction and intends to backtrack. 
However, after completing a sing-step backtracking at step 8, it again decides to go back to the wrong place it was at step 7. 
This behavior of oscillating between two locations persists until navigation failure at step 15.
On the other hand, the global planning space enables the agent to capture global environment layouts and successfully correct previous wrong decisions. 
At step 4, the agent also starts by navigating in the wrong direction, just like the local planning variants. 
But the predicted long-term goal effectively guides it back on the right track at step 8, concluding with successful navigation.

As shown in Fig.~\ref{fig:vis2}, the practical sliding-forbidden steps can cause the agent to get stuck in obstacles and lead to navigation failure.
For instance, in the absence of Tryout, the agent is unable to proceed forward once its chassis collides with the wall (at steps 6 and 7).
This situation persists until the end of navigation (at step 14), where the agent does not succeed in escaping the deadlock, ultimately leading to navigation failure. 
Conversely, the integration of Tryout control in our model effectively addresses this issue. 
At step 4, Tryout is triggered upon colliding with the wall, causing the agent to twist and stagger away from the obstacle. 
This helps to navigate around the obstacle, with the navigation accomplished at step 6 successfully.

\section{Conclusion}\label{sec:conclusion}
In summary, this paper introduces ETPNav, a novel navigation system that leverages topological maps for VLN-CE.
We first propose an online mapping method via waypoint self-organization to enable robust long-range planning of an agent. 
This scheme doesn't require any prior environmental experience and satisfies the demand of navigation in realistic scenarios. 
Then, we systematically examine our topo map's key design choices and empirically show that a concise depth-only design can be optimal for waypoint prediction. 
Furthermore, we address an often-neglected issue in VLN-CE - obstacle avoidance, with a simple and effective trial-and-error controller. 
Extensive experiments demonstrate the effectiveness of the proposed method, yielding more than 10\% and 20\% absolute improvements over prior state-of-the-art on R2R-CE and RxR-CE benchmarks, respectively. 
We hope this work can serve as a strong baseline for further research on this challenging task. 

\nbf{Future Work.}
ETPNav employs pose-reading sensors to facilitate topological mapping. 
However, in real-world applications, the presence of sensor actuation noise may introduce complexities into the mapping process. 
We take the noise analysis and robustness study in such scenarios as the future work plan.


\ifCLASSOPTIONcaptionsoff
  \newpage
\fi

\bibliographystyle{IEEEtran}
\bibliography{APC}

\end{document}